\documentclass{article}

\usepackage[main, final]{neurips/neurips_2026}

\usepackage[utf8]{inputenc} % allow utf-8 input
\usepackage[T1]{fontenc}    % use 8-bit T1 fonts
\usepackage{hyperref}       % hyperlinks
\usepackage{url}            % simple URL typesetting
\usepackage{booktabs}       % professional-quality tables
\usepackage{amsfonts}       % blackboard math symbols
\usepackage{nicefrac}       % compact symbols for 1/2, etc.
\usepackage{microtype}      % microtypography
\usepackage{xcolor}         % colors
\usepackage{enumitem}

\usepackage{soul}

\usepackage{makecell}
\usepackage[breakable,skins]{tcolorbox}

\usepackage{amsmath}
\usepackage{amssymb}
\usepackage{mathtools}
\usepackage{amsthm}
\usepackage{dsfont}

\usepackage{multirow}
\usepackage{algorithm}
\usepackage{algpseudocode}

\usepackage{graphicx}
\usepackage{subcaption}
\usepackage{wrapfig}

\usepackage[capitalize,noabbrev]{cleveref}

\newcommand{\agentdecision}{z}
\newcommand{\agentdecisionRV}{Z}
\newcommand{\agentdecisionsp}{\mathcal{Z}}
\newcommand{\decision}{d}
\newcommand{\decisionRV}{D}
\newcommand{\decisionsp}{\mathcal{D}}

\newcommand{\supervisorinfo}{t}
\newcommand{\supervisorinfoRV}{T}
\newcommand{\supervisorinfosp}{\mathcal{T}}

\newcommand{\latentsignal}{\mathbf{s}}
\newcommand{\latentsignalRV}{\mathbf{S}}
\newcommand{\latentsignalsp}{\mathcal{S}}

\newcommand{\payoffstate}{y}
\newcommand{\payoffstateRV}{Y}
\newcommand{\payoffstatesp}{\mathcal{Y}}

\newcommand{\scoringrule}{U}

\newcommand{\agentfeature}{x}
\newcommand{\agentfeatureRV}{X}
\newcommand{\agentfeaturesp}{\mathcal{X}}

\newcommand{\dgp}{\pi}
\newcommand{\distover}[1]{\Delta(#1)}

\newcommand{\expect}[2]{\E_{#1}\left[#2\right]}

\newcommand{\rational}{\mathcal{V}}

\newcommand{\llm}{\texttt{LLM}}
\DeclareMathOperator*{\E}{\mathbb{E}}

\DeclareMathOperator*{\argmax}{argmax}

\newcommand{\reward}{R}

\newcommand{\sysname}{\textsc{ComplLLM}}
\title{\sysname{}: Fine-tuning LLMs to Discover Complementary Signals for Decision-making}

\author{%
  Ziyang Guo\\
  Northwestern University\\
  Evanston, IL \\
  \texttt{ziyang.guo@northwestern.edu} \\
  \And
  Yifan Wu \\
  Microsoft Research \\
  Cambridge, MA\\
  \texttt{yifan.wu2357@gmail.com} \\
  \AND
  Jason Hartline \\
  Northwestern University\\
  Evanston, IL \\
  \texttt{hartline@northwestern.edu} \\
  \And
  Kenneth Holstein \\
  EPFL \\
  Lausanne, Switzerland\\
  \texttt{ken.holstein@epfl.ch} \\
  \And
  Jessica Hullman \\
  Northwestern University\\
  Evanston, IL \\
  \texttt{jhullman@northwestern.edu} \\
}

\newcommand{\mvspace}[1]{\vspace{#1}}

\begin{document}

\maketitle

\begin{abstract}
Multi-agent decision pipelines can outperform single agent workflows when \textit{complementarity} holds, i.e., different agents bring unique information to inform a final decision.
We propose \sysname{}, a post-training framework based on decision theory that fine-tunes a decision-assistant LLM to output signals that complement existing agent decisions, using complementary information as reward.
We validate \sysname{} on synthetic and real-world tasks involving domain experts, demonstrating how the approach recovers known complementary information, produces plausible explanations of complementary signals to support downstream decision-makers, and improves human and LLM agent decision performance in controlled studies.
\end{abstract}

\section{Introduction}

Multi-agent collaboration is increasingly adopted in complex decision workflows.
For example, a clinician or AI decision agent may consult a computer vision model and a written report from a radiologist to decide whether to order a biopsy for the patient. A content moderator decides whether to approve a post based on reviews from different human raters. A program chair decides on paper acceptance based on reviewer comments and automated assessments from an LLM. 
In such settings, the central challenge for the downstream decision-maker is how to integrate information from different sources. 
The decision-maker must determine which parts of the available information are relevant to the task, which parts are already reflected in an upstream agent's recommendation, and which parts provide complementary evidence that could improve the final decision. This requires sufficient attention to identify relevant information embedded in long unstructured texts, as well as domain expertise and statistical reasoning to determine which signals are not redundant with other sources like existing agent decisions.

%These settings motivate the extraction of the decision signals that target the intergration problem.
Instead of leaving a decision-maker to reason over all available unstructured information on their own, a promising idea is to surface a more compact set of signals, optimized to present the most decision-relevant information that is also complementary in light of existing agent recommendations.
However, most machine-learning interpretability methods are not designed to address complementarities nor decision-making per se~\citep{guo2026explanations}. 
Explanations typically characterize why a single model produced its output, often with respect to that model's own inputs (e.g., feature attributions, saliency, rationales).
In collaborative workflows, the bottleneck is often different.
For a clinician using a vision model in diagnosis, for example, what is often needed is not simply a restatement of the model's reasoning, but a list of features of patient information that are not considered by existing agent recommendations. %   to guide their decision together with the model score.

Our work formalizes available-but-unintegrated evidence as \textbf{complementary signals}.
% , defined as a set of discrete, interpretable findings extracted from one agent's information that capture decision-relevant information beyond the other agent's recommendation. 
We consider settings with two ``agents'': an upstream agent that produces a recommendation $\agentdecisionRV$ (e.g., a vision model's risk score on an X-ray image), and a supervisor agent that has access to potentially complementary unstructured information $\supervisorinfoRV$ (e.g., a text such as a radiology report). 
% \ken{The word ``additional'' here might be read as implying that the supervisor agent has access to a superset of the information observed by the upstream agent. Consider changing the wording to something like ``a supervisor agent that has access to different information'' or ``a supervisor agent that has access to potentially complementary unstructured information''} 
The goal is to identify a set of discrete, interpretable \textit{signals} (i.e., findings) $\latentsignalRV$ in $\supervisorinfoRV$ that capture complementary decision-relevant information not already conveyed by $\agentdecisionRV$, i.e., $\latentsignalRV$ that can improve the best-attainable decision conditioned on $\agentdecisionRV$.
%These signals can be used as a principled object for both interpretation (what to look at) and evaluation (how much value remains beyond the recommendation). \jessica{can omit prev sentence for space if needed}
% The goal is to identify signals in $\supervisorinfoRV$ that (i) are decision-relevant, and (ii) provide incremental value over what is already encoded in $\agentdecisionRV$ conditioned on the distribution of state learned from historical data.

%Building on this formulation, 
We propose \sysname{}, a framework that (1) defines complementary value using the best-attainable performance on the decision problem, and (2) trains a language model to act as a complementary signal extractor from unstructured text. 
Formally, \sysname{} learns a mapping from unstructured textual information ($\supervisorinfoRV$) and recommendations ($\agentdecisionRV$) to a set of structured signals, $\llm: \supervisorinfosp \times \agentdecisionsp \rightarrow \latentsignalsp$, which prioritizes signals that meaningfully improve best-attainable decision quality relative to using the recommendation alone, rather than signals that are merely frequent or important.
This shifts the role of ``explanation'' from justifying the agent recommendation to surfacing actionable complementary information that a supervisor should consider precisely because it is not already reflected in $\agentdecisionRV$.
\sysname{} can be used for cases where the two agents are distinct, or where they represent the same human or model, i.e., an agent makes an initial decision using its available information, and then uses an LLM to do a second pass to surface overlooked cues in that same information.
%In this self-complementation setting, the LLM is not introducing new data; it is helping the agent recover decision-relevant information that was present but under-exploited. 
%Hence, the framework synthesizes ``multi-agent collaboration'' (different agents, different information) and ``self-auditing/self-correction'' (one agent, same information, improved extraction and utilization).

We validate \sysname{} across multiple domains. 
We first use a synthetic setting where complementary signals are controlled by construction to test the recoverability of the complementary signals. 
We then demonstrate use of the framework on three real-world decision-making tasks: identifying complementary signals in radiology reports that improve upon a vision model's recommendation on cardiac dysfunction; identifying the signals that are more or less focal to a specific group compared to the average human annotator in content moderation; and identifying the information that an LLM-authored review misses in human-written reviews.
We evaluate the effectiveness of the signals generated by \sysname{} by a preregistered within-subject human study for content moderation, a qualitative interview with two medical domain experts (one cardiologist and one internist) for radiology diagnosis, and an LLM experiment (Gemini 2.0 Flash) for paper reviewing.

\section{Related Work}

\subsection{Topic \& Hypothesis Generation}

\begin{figure}
    \centering
    \includegraphics[width=0.5\linewidth]{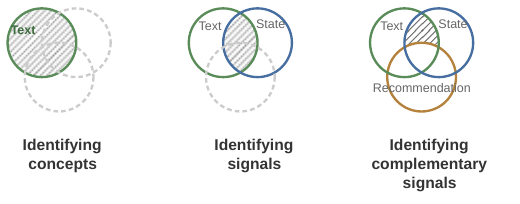}
    \mvspace{-3mm}
    \caption{A graphical representation of how our work goes beyond simply identifying concepts that appear in text or signals that are predictive of a target state, by also ensuring that signals complement the existing recommendation. The diagonal stripes (twill pattern) represent the target of the corresponding methods.}
    \mvspace{-3mm}
\label{fig:concept_signal_compl_signal}
\end{figure}

A large literature studies topic (or \textit{concept}) generation as a way to summarize or organize corpora: classical probabilistic models such as LDA infer latent topics as word distributions~\citep{blei2003latent}, while neural and embedding-based variants improve coherence and interpretability (e.g., ProdLDA/AVITM~\citep{srivastava2017autoencoding}, Embedded Topic Model (ETM)~\citep{dieng2020topic}, and other neural variational topic models such as Miao et al.~\citep{miao2017discovering}).
Recent representation-first pipelines treat topic modeling as clustering in embedding space (e.g., Top2Vec~\citep{angelov2020top2vec}, BERTopic~\citep{grootendorst2022bertopic}, and newer embedding-centric formulations~\citep{angelov2024topic}), and LLMs increasingly support prompt-based topic generation/labeling with greater steerability (e.g., TopicGPT~\citep{pham2023topicgpt}).
These methods primarily target concepts or topics (i.e., patterns grounded in text) without necessarily requiring a link to a target state.

A parallel line of work targets hypothesis (or \textit{signal}) generation, which produces interpretable candidate factors that are grounded in text and also correlated with a state of interest (\Cref{fig:concept_signal_compl_signal}).
Examples include using SAEs to generate hypotheses from latent features with LLM interpretations (e.g., HypotheSAEs~\citep{movva2025sparse}), learning natural-language descriptions of distributional differences and using them in goal-driven discovery \citep{zhong2024explaining, zhou2024hypothesis}, and using LLM-proposed differences followed by statistical validation for causal inference on text-derived outcomes~\citep{modarressi2025causal}.
As shown in \Cref{fig:concept_signal_compl_signal}, our work shifts focus to identify complementary signals, i.e., signals grounded in the supervisor information that add incremental decision value relative to an existing recommendation (not merely frequent, salient, or globally predictive signals).

\subsection{Human-AI Decision-making}
A growing body of work studies AI-assisted human decision-making based on its importance for legal and ethical accountability~\citep{bo2021toward, boskemper2022measuring, bondi2022role, schemmer2022meta}.
A recent meta-analysis \citep{vaccaro2024combinations} finds that, on average, human-AI teams perform worse than the better of the two agents alone. 
A growing body of work seeks to evaluate and enhance complementarity in human-AI systems \citep{bansal2019updates,bansal2021does, bansal2021most,wilder2021learning,hemmer2022effect,holstein2023toward,rastogi2023taxonomy, mozannar2024effective,guo2025value}. 
Some approaches explicitly incorporate human expertise in developing machine learning models or human-AI collaboration pipelines, such as by learning to defer~\citep{mozannar2024show,raghu2019algorithmic, keswani2022designing,keswani2021towards, okati2021differentiable,chen2022machine}.
Others develop alternative algorithms, e.g., with provable guarantees, that exploit cases where humans have additional contextual knowledge~\citep{alur2024distinguishing,corvelo2023human,straitouri2023improving,de2024towards,arnaiz2025towards}. 
Closest to our work, Guo et al.~\citep{guo2025value} provide a framework for assessing the complementary information value of arbitrary signals in a decision context. We demonstrate using complementary information as a fine-tuning objective, enabling LLM-based explanations of unexploited signals in unstructured text available at decision time. 

%\ziyang{econ \& HCI}

\section{Theoretical Framework}

\begin{figure*}[t]
    \centering
    \includegraphics[width=0.8\textwidth]{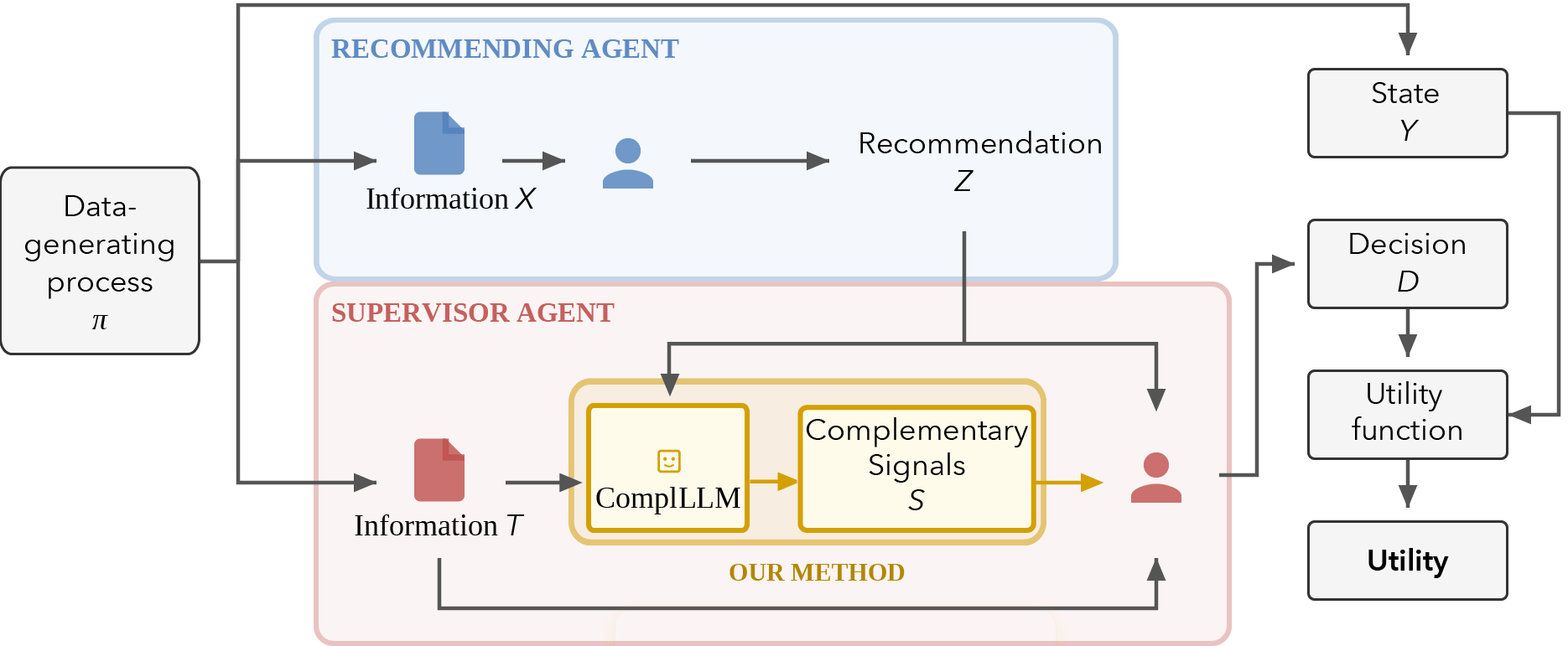}
    % \mvspace{-4mm}
    \caption{The ``two agents'' setting in our framework.
    % (which can be generalized to settings with more than two agents)
    The recommending agent makes a recommendation $\agentdecisionRV$ based on their own features $\agentfeatureRV$, and the supervisor agent aggregates their own information $\supervisorinfoRV$ with $\agentdecisionRV$ to make a final decision $\decisionRV$.}
    \mvspace{-3mm}
    \label{fig:dag}
\end{figure*}

% In this section, we first define the setting of the decision-making task and then define our optimization problem of complementary signals. \jessica{whose optimization problem? writing is very hard to follow}

We consider a decision-making task with an upstream, recommending agent and a downstream, supervisor agent where, for each realization of the process, the recommending agent makes a recommendation $\agentdecisionRV \in \agentdecisionsp$ based on their own features $\agentfeatureRV \in \agentfeaturesp$, and the supervisor aggregates their information $\supervisorinfoRV \in \supervisorinfosp$ with $\agentdecisionRV$ to make a final decision $\decisionRV \in \decisionsp$.
% We assume that the supervisor's information $\supervisorinfoRV$ is generated by a set of latent signals $\latentsignalRV^* \in \latentsignalsp$.
The goal of the supervisor is to determine whether there is more information in $\supervisorinfoRV$--in the form of a set of signals $\latentsignalRV \in \latentsignalsp$--over $\agentdecisionRV$ that can improve their utility.

We do not assume any relationship between the recommending agent's feature representation $\agentfeatureRV$ and the supervisor's information $\supervisorinfoRV$.
For example, in the radiology diagnosis task, the features of the recommending agent (vision model) $\agentfeatureRV$ are the X-ray image, and the features of the supervisor (clinician) $\supervisorinfoRV$ are the radiology report and the patient's medical history.
In the content moderation task, the features of the recommending agent (a specific group of human annotators) $\agentfeatureRV$ and the features of the supervisor (moderator) $\supervisorinfoRV$ are both the text content.
% Therefore, the framework can be used to identify both the information that is missing in $\agentfeatureRV$ but is available in $\supervisorinfoRV$, and the information that is available in both $\agentfeatureRV$ and $\supervisorinfoRV$ but is not elicited by $\agentdecisionRV$.
% \jessica{somewhere above we need to motivate why this captures actual decision problems people are about}

We use decision theory to characterize the decision-making process as a \textit{decision problem} and an \textit{information structure}.
The decision problem consists of three components: the state space $\payoffstatesp$, the decision space $\decisionsp$, and the utility function $\scoringrule: \decisionsp \times \payoffstatesp \rightarrow \mathbb{R}$.
The state $\payoffstate \in \payoffstatesp$ represents the underlying state of the world that is relevant to the supervisor's utility.
Since the utility function $\scoringrule$ implicitly identifies the other two components $\payoffstatesp$ and $\decisionsp$, henceforth, we will use $\scoringrule$ to denote a decision problem.

The information structure is a joint distribution over the space of features of the recommending agent $\agentfeaturesp$, the features of the supervisor $\supervisorinfosp$, and the state $\payoffstatesp$, denoted as $\dgp \in \distover{\agentfeaturesp \times \supervisorinfosp \times \payoffstatesp}$.
Given a set of observations $\{(\agentfeature_i, \supervisorinfo_i, \payoffstate_i)\}_{i\in [N]}$, 
we can estimate $\dgp$ on recommendation $\agentdecisionRV$ and derived discrete signals $\latentsignalRV$, assuming that the state is discrete, e.g., $\payoffstateRV \in \{0, 1\}$.
We denote the probability of observing $\agentdecisionRV=\agentdecision$, $\latentsignalRV=\latentsignal$, and $\payoffstateRV=\payoffstate$ as $\dgp(\agentdecision, \latentsignal, \payoffstate)$.
% The latent signals $\latentsignal \in \latentsignalsp$ are the underlying conditions that generate the supervisor's information $\supervisorinfo$, i.e., $\supervisorinfo = f_{\supervisorinfo}(\latentsignal)$.
We denote the set of all possible signals as a combination of $M$ basic signals, i.e., $\latentsignalsp = \{0, 1\}^M$.
Each of the basic signals is a binary random variable, indicating the presence or absence of a certain signal, e.g., \Cref{tab:example_notation}.
We use upper-case letters to represent the random variables, e.g., $\latentsignalRV$, $\agentdecisionRV$, $\payoffstateRV$, and lower-case letters to represent a value, i.e., $\latentsignal$, $\agentdecision$, $\payoffstate$.
We denote the $j$-th basic signal as $\latentsignalRV_j \in \{0, 1\}$, and its absence or presence as $\latentsignal_j \in \{0, 1\}$.
% Hence, a latent signal $\latentsignalRV$ can be represented as a combination of $m$ basic signals, i.e., $\latentsignalRV = (\latentsignalbasicRV_{j_1}, \latentsignalbasicRV_{j_2}, \cdots, \latentsignalbasicRV_{j_m})$, where $j_1, \cdots, j_m \in \{1, 2, \cdots, M\}$.
% We define $\latentsignalsp^{(k)}$ to denote the signal space where there is at most $k$ basic signals that are present at the same time, i.e., for any $(\latentsignalbasic_{j_1}, \latentsignalbasic_{j_2}, \cdots, \latentsignalbasic_{j_m}) \in \latentsignalsp^{(k)}$, $\sum_{i=1}^m \latentsignalbasic_{j_i} \le k$.

% a latent signal with $m$ basic signals is $\latentsignalRV = (\latentsignalbasicRV_{j_1}, \latentsignalbasicRV_{j_2}, \cdots, \latentsignalbasicRV_{j_m})$, and lower-case letters to represent a value, i.e., $\latentsignal = (\latentsignalbasic_{j_1}, \latentsignalbasic_{j_2}, \cdots, \latentsignalbasic_{j_m})$
% \footnote{$j_1, \cdots, j_m \in \{1, 2, \cdots, k\}$ is the index of the basic signal in the full set.}.
% The corresponding space of all possible latent signals is $\latentsignalsp_{j_1} \times \latentsignalsp_{j_2} \times \cdots \times \latentsignalsp_{j_m}$.

The value of a signal is defined as the improvement that can be attained in the best attainable performance when that signal is provided.
% For notational simplicity, we write the value of signals as best-responding payoff $\hat{\scoringrule}$ to a distribution. \jessica{need to make this more concrete - best responding payoff makes it very abstract. Make clear you are talking about identifying the best attainable payoff for the decision task} 
Concretely, a signal realization $\latentsignal$ induces posterior distribution $\dgp(\payoffstate | \latentsignal)$. We define $\hat{\scoringrule}$ as the payoff of the decision that best-responds to the posterior distribution on the realization $\payoffstateRV=\payoffstate$.
% \mvspace{-2mm}
\begin{equation*}
% \hat{\scoringrule}(\dgp(\payoffstate | S)) = 
% \max_{\decision \in \decisionsp} \expect{\payoffstate'\sim \dgp(\payoffstate | S)}{\scoringrule(\decision, \payoffstate')}.
    \hat{\scoringrule}(\dgp(\cdot | \latentsignal), \payoffstate) := \scoringrule(\argmax_{\decision \in \decisionsp} \expect{\payoffstate'\sim \dgp(\cdot | \latentsignal)}{\scoringrule(\decision, \payoffstate')}, \payoffstate)
\end{equation*}
\mvspace{-3mm}

The best attainable performance given $\latentsignalRV$ is the expected payoff of the best-responding payoff $\hat{\scoringrule}$: $\E_{\latentsignal, \payoffstate\sim\dgp}\left[\hat{\scoringrule}(\dgp(\cdot|\latentsignal), \payoffstate)\right]$.
% We denote the language model that extracts the signals as $\llm(\cdot,\cdot): \supervisorinfosp \times \agentdecisionsp \rightarrow \latentsignalsp$.
The \textbf{complementary value} of the signal extracted by a language model $\llm(\cdot,\cdot): \supervisorinfosp \times \agentdecisionsp \rightarrow \latentsignalsp$ in a decision problem $\scoringrule$ is the difference between the expected best attainable payoff
%\footnote{A scoring rule $\hat{\scoringrule}: \distover{\payoffstatesp} \times \payoffstatesp \rightarrow \mathbb{R}$ is proper if truthful reporting of the payoff state $\payoffstate$ maximizes the expected utility for the decision-maker, i.e., $\expect{\payoffstate \sim p}{\hat{\scoringrule}(p, \payoffstate)} \geq \expect{\payoffstate \sim p}{\hat{\scoringrule}(q, \payoffstate)}$ for any $p, q \in \distover{\payoffstatesp}$.} 
on observing $\llm(\supervisorinfoRV, \agentdecisionRV)$ and $\agentdecisionRV$, and the expected best attainable payoff on observing $\agentdecisionRV$: 
% \jessica{best responding payoff is awkward, spell out what this means. when we write like we expect readers to be theorists our papers tend to get rejected.}
\begin{equation}
    % \begin{aligned}
        \label{eq:rational}
        \rational^{\scoringrule, \agentdecisionRV}(\llm) :=  \expect{\supervisorinfo, \agentdecision, \payoffstate \sim \dgp}{\hat{\scoringrule}(\dgp(\cdot | \llm(\supervisorinfo, \agentdecision), \agentdecision), \payoffstate)} - \expect{\agentdecision, \payoffstate \sim \dgp}{\hat{\scoringrule}(\dgp(\cdot | \agentdecision), \payoffstate)}
    % \end{aligned}
    % \text{s.t. } & \latentsignalRV = (\latentsignalbasicRV_{j_1}, \latentsignalbasicRV_{j_2}, \cdots, \latentsignalbasicRV_{j_m}) \\
    % & \latentsignalRV \in \latentsignalsp^{(k)}
    % - \\ \expect{\latentsignal}{\max_{\decision \in \decisionsp} \expect{\payoffstate}{\scoringrule(\decision, \payoffstate) \mid \latentsignalRV = \latentsignal}}
\end{equation}
\mvspace{-4mm}
% where $\dgp(\payoffstate | \cdot)$ is the posterior distribution of $\hat{\dgp}$ over $\payoffstate$.

We demonstrate our framework in \Cref{tab:example_notation}, using a radiology diagnosis task. 

%     The state space is $\payoffstatesp = \{0, 1\}$ for the presence of cardiac dysfunction. 
% The decision space $\decisionsp = [0, 1]$ is the same as the prediction space. Both the utility function and the best-responding utility function are the Brier loss of the prediction $\hat{\scoringrule}(p, \payoffstate) = 1- (p - \payoffstate)^2$. The space of signals are the presence of symptoms in textual clinical notes, for example, $\latentsignalRV_1 = 1$ represents the enlargement of cardiac silhouette, $\latentsignalRV_2 = 1$ represents the presence of pleural abnormalities, etc. 

% In preamble (optional but recommended):
% \usepackage{booktabs}

\begin{table}[t]
\centering
\small
\setlength{\tabcolsep}{4pt} % tighter spacing for double column
\caption{Notation summary (instantiated with the example of radiology-diagnosis).}
\renewcommand{\arraystretch}{1.05}
\begin{tabular}{@{}l p{0.9\columnwidth}@{}}
\toprule
\textbf{Notation} & \textbf{Radiology Diagnosis instantiation} \\
\midrule
$\payoffstateRV$ & Payoff state $\in\{0,1\}$ for the presence of cardiac dysfunction. \\
$\agentdecisionRV$ & Prediction from the vision model on the probability of the presence of cardiac dysfunction $\in[0,1]$. \\
$\agentfeatureRV$ & Features of the vision model (X-ray image). \\
$\supervisorinfoRV$ & Features of the clinician (radiology report and patient's medical history). \\
$\decisionRV$ & Final decision by the clinician $\in\{0,1\}$ (e.g., whether to order a biopsy). \\
\hline
\multicolumn{2}{l}{Signals in $\supervisorinfoRV$}\\
$\latentsignalRV_1$ & presence of enlarged cardiac silhouette $\in \{0, 1\}$\\
$\latentsignalRV_2$ & presence of pleural abnormalities $\in \{0, 1\}$. \\
\dots & \\
\bottomrule
\end{tabular}
\mvspace{-3mm}
\label{tab:example_notation}
\end{table}

\mvspace{-2mm}
\section{Methods}
\mvspace{-2mm}

% In the previous section, we established theoretically that the discovery of complementary signals for all the decision problems is equivalent to the discovery of complementary signals for a proper scoring rule.
% We now present our method to train an LLM agent to identify the these signals for a decision problem in practice. \jessica{this sentence is not conveying any new information}

% Given training data $\{(t_i,z_i,y_i)\}_{i=1}^N$ and a decision problem $U$, \sysname{} fine-tunes an LLM to extract signals $s_i=LLM_\theta(t_i,z_i)$ that maximize complementary value over $z_i$. The training pipeline has three stages. First, we estimate a reference model over candidate signals, recommendations, and states by prompting an LLM to propose signal types and annotate their occurrence. Second, we use the reference model to construct SFT targets: the minimal supported signal set that maximizes the improvement in best-attainable payoff over the recommendation. Third, we further optimize the model with GRPO using a reward proportional to the complementary payoff gain of supported signals and zero reward for unsupported signals.

Given a training dataset $\{(\supervisorinfo_i, \agentdecision_i, \payoffstate_i)\}_{i\in [N]}$ and a decision problem $\scoringrule$, we fine-tune a large language model $\llm_\theta$to extract a set of signals $\latentsignal_i = \llm_{\theta}(\supervisorinfo_i, \agentdecision_i)$ that maximizes the complementary value over $\agentdecision_i$, as defined in \Cref{eq:rational}. 
% Intuitively, the learned representation $\latentsignalRV$ should encode the decision-relevant information contained in the supervisor's input $\supervisorinfoRV$ that is either missing from or underutilized by the agent's recommendation $\agentdecisionRV$.
First, we estimate the data-generating process as the joint distribution between the signals, agent decision, and the state. We use this estimated distribution as the reference model to define the complementary value.
Second, we use the reference model to construct SFT targets, which is defined as the minimal supported signal set that maximizes the improvement in best-attainable payoff over the recommendation.
Last, we further optimize the model with GRPO using a reward proportional to the complementary information value of supported signals.

\mvspace{-2mm}
\subsection{Estimating the Data-Generating Process}
\mvspace{-2mm}

% The estimation of the data-generating process, $\dgp \in \distover{\latentsignalsp\times \agentdecisionsp\times \payoffstatesp}$, contains three steps: \jessica{might be confusing to have three steps followed by three steps. Also you're using passive voice a lot, which makes it seem less exciting} identify the space of signals, identify \jessica{which signals are present for each instance} the presence of the signals in each instance, and estimate the joint distribution.

We estimate the data-generating process by prompting an LLM model in two rounds.
In the first round, we identify the space of possible signals, denoted as $\latentsignalsp = \{0, 1\}^M$, and in the second round, we identify whether a signal occurs in each instance $(\supervisorinfo_i, \agentdecision_i, \payoffstate_i)$, denoted as $\{\latentsignal_i^{(0)}: \latentsignal_i^{(0)} \in \latentsignalsp\}_{i\in[N]}$.

% We estimate the data-generating process using empirical samples from the training dataset and an LLM-based signal extractor.
We identify the space of signals by initializing a set of signals on each instance.
We use an LLM reference model $\llm_{\text{ref}}$ to output the signals that occur in the supervisor's information $\supervisorinfo_i$.
The union of the signals across all instances forms the space of all possible signals in our estimate, $\latentsignalsp = \{0, 1\}^M$, where $M$ is the number of signals in the union set.
To improve stability, we sample $\zeta = 7$ outputs at temperature 0.7 and only keep the signals with frequency larger than $N_\tau * \zeta$~\footnote{We use $N_{\tau} = \frac{z_{1-\delta/2}^2 p (1-p)}{\epsilon^2}$ as a heuristic minimum sample size, motivated by the normal approximation for a binomial proportion. Here $p = \dgp(\payoffstateRV = 1)$ is the marginal prevalence and $z_{1-\delta/2}$ is the $1-\delta/2$ quantile of the standard normal distribution. This heuristic filters infrequent signals to reduce estimation noise.}.

In the second round, for each signal $j \in [M]$ and each instance $i \in [N]$,  we prompt $\llm_{\text{ref}}$ again to only output whether signal $j$ occurs in $\supervisorinfo_i$.
% We denote the initial signal presence vector for $\supervisorinfo_i$ as $\latentsignal_i^{(0)} \in \{0, 1\}^M$ and $\latentsignal_{i, j}^{(0)} = 1$ if the $j$-th basic signal is present in the supervisor's information $\supervisorinfo_i$.
% The completion length is set to 2,048 tokens and the temperature is set to $0.7$.
We estimate the posterior distribution $\dgp(\payoffstate| \cdot)$ by a regression model (\Cref{alg:greedy_adaptive}) using $\{\latentsignal^{(0)}_{i}, \payoffstate_i\}_{i\in[N]}$, which we subsequently use to generate training data for SFT and the reward function for RL.
\subsection{Supervised Fine-tuning (SFT) with Generated Complementary Signals}
\label{sec:sft}
% \mvspace{-2mm}
\paragraph{Generating Complementary Signals.}
We generate the training data for SFT as the complementary signals $\latentsignal_i^{(1)} \in \{0, 1\}^M$ for each instance $(\supervisorinfo_i, \agentdecision_i, \payoffstate_i)$.
Given $\latentsignal_i^{(0)}$ denoting the occurrence of basic signals in $\supervisorinfo_i$, we generate $\latentsignal_i^{(1)}$ by selecting the minimal set of basic signals that both occur and maximize the improved best-attainable payoff over the agent decision  $\agentdecision_i$ on instance $i$ (which we require to be more than a minimum threshold $\epsilon$), as these signals convey decision-relevant information that is unexploited by $\agentdecision_i$.
% \begin{equation}
%     \begin{aligned}
%         & \latentsignal_i^{(1)} = (\latentsignal_{i, 1}^{(1)}, \ldots, \latentsignal_{i, M}^{(1)}) \\
%         & \text{ s.t. } \latentsignal_{i, j}^{(0)} = 1 \text{ for any } \latentsignal_{i, j}^{(1)} = 1,\text{ and} \\
%         & \quad \quad \hat{\scoringrule}(\dgp(\payoffstate|\latentsignal_i^{(1)},\agentdecision_i), \payoffstate_i) > \hat{\scoringrule}(\dgp(\payoffstate|\agentdecision_i), \payoffstate_i) + \epsilon
%     \end{aligned}
% \end{equation}
\begin{align*}
&\min_{\latentsignal_i^{(1)}}|\latentsignal_i^{(1)}|\\
s.t. \ & \latentsignal_i^{(1)} \in \argmax_{\text{supp}(\latentsignal) \subseteq \text{supp}(\latentsignal_{i}^{(0)})} \left[\hat{\scoringrule}(\dgp(\cdot|\latentsignal, \agentdecision_i), \payoffstate_i) - \hat{\scoringrule}(\dgp(\cdot|\agentdecision_i), \payoffstate_i)\right]\text{, and}\\
& \hat{\scoringrule}(\dgp(\cdot|\latentsignal_i^{(1)}, \agentdecision_i), \payoffstate_i) > \hat{\scoringrule}(\dgp(\cdot|\agentdecision_i), \payoffstate_i) + \epsilon
\end{align*}
where $\text{supp}(\mathbf{s})=\{j: \mathbf{s}_j=1\}$.
If there is no solution (i.e., no signals improve over the agent decision more than $\epsilon$), we set $\mathbf{s}_i^{(1)}$ as empty (i.e., no complementary signals identified).

\mvspace{-2mm}
\paragraph{Supervised Fine-tuning.}
We fine-tune $\llm_{\theta}$ using the generated complementary signals $\latentsignal_i^{(1)}$ as ground-truth labels.
To improve the efficiency of this process \citep{wang2025octothinker, gandhi2025cognitive}, we use $\llm_\text{ref}$ to generate a Chain-of-Thought (CoT) given the supervisor's information $\supervisorinfo_i$, the agent's decision $\agentdecision_i$, and the ground truth of $\latentsignal_i^{(1)}$.
We require the format of the CoT to identify the following: evidence for signals, the relevance to the state, and the complementary value relevant to the agent decision.
% We compose the thinking content and the generated complementary signals to get the training data in supervised fine-tuning. \jessica{awkwardly worded}

\mvspace{-2mm}
\subsection{Reinforcement Learning}
% \mvspace{-2mm}

We sequentially fine-tune $\llm_{\theta}$ after SFT using \textbf{Group Relative Policy Optimization (GRPO)} \cite{shao2024deepseekmath} to maximize the objective given in \Cref{eq:rational}. % in addition to the supervised fine-tuning.
% By \Cref{thm:lattice}, $\latentsignalRV$ is also improved over other decision problems $\scoringrule'$ that can be derived from $\scoringrule$ with swap $g(\cdot)$, scaling $w(\cdot)$, and a constant shift $C$.
\mvspace{-2mm}
\paragraph{Reward function.}
We design the reward function using the best-attainable payoff.
The reward function $\reward: \latentsignalsp \times \agentdecisionsp \times \payoffstatesp \rightarrow \mathbb{R}$ rewards signals only when they increase best-attainable payoff beyond the agent's recommendation, i.e., when they provide complementary information value.
We only score signals that are \textit{supported} by the supervisor's information to prevent rewarding hallucinations, where $\latentsignal$ is supported if every basic signal in $\latentsignal$ occurs in the supervisor's information, i.e., $\latentsignal_{i, j}^{(0)} = 1$ for all $j \in [M]$ such that $\latentsignal_{j} = 1$. 
\mvspace{-2mm}
% to incentivize $\llm_{\theta}$ to extract signals $\latentsignal$ that contain more complementary value over the agent's decision $\agentdecision$ on instance $(\supervisorinfo, \agentdecision, \payoffstate)$.
\begin{equation}
    \label{eq:reward}
    % \begin{aligned}
    \reward(\latentsignal, \agentdecision_i, \payoffstate_i) = 
    % &\sum_{\substack{j \in [M] \\ \text{ s.t. } \\ \latentsignal_{j} = 1}} 
     \begin{cases}
        1 \qquad \,\, & \text{if both $\latentsignal$ and $\latentsignal^{(1)}_i$ are empty}\\
        (\hat{\scoringrule}(\dgp(\cdot|\latentsignal,\agentdecision_i), \payoffstate_i)- \hat{\scoringrule}(\dgp(\cdot|\agentdecision_i), \payoffstate_i))/\alpha_i \,\, & \text{if $\latentsignal$ is supported and $\latentsignal^{(1)}_i$ is not empty}  \\
        0  \qquad \,\, & \text{else}
    \end{cases}
    % \end{aligned}
\end{equation}
\mvspace{-4mm}

where $\alpha_i=\hat{\scoringrule}(\dgp(\cdot|\latentsignal_i^{(1)},\agentdecision_i), \payoffstate_i)- \hat{\scoringrule}(\dgp(\cdot|\agentdecision_i), \payoffstate_i)$ is the best attainable payoff on instance $i$ with the identified complementary signals in SFT training data, which we use as a normalizer.
\mvspace{-2mm}
\paragraph{Group Relative Policy Optimization (GRPO).}

We fine-tune $\llm_{\theta}$ using the GRPO algorithm \citep{shao2024deepseekmath}.
At each training step, the model generates $G$ candidate signals $\latentsignal_{i}^1, \cdots, \latentsignal_{i}^G$ for each instance $i$.
The advantage function of GRPO is defined as the reward difference between the candidate signal and the average reward of the candidates.
\mvspace{-2mm}
\begin{equation}
    \begin{aligned}
        A_j = \reward(\latentsignal_{i}^j, \agentdecision_i, \payoffstate_i) - \frac{1}{G} \sum_{k=1}^G \reward(\latentsignal_{i}^k, \agentdecision_i, \payoffstate_i)
    \end{aligned}
\end{equation}
\mvspace{-4mm}

With the above advantage function, we train $\llm_{\theta}$ using the objective function defined in \cite{shao2024deepseekmath}.

% \subsection{Iterating the Training Process}

% We iterate the training process between estimating the data-generating process and fine-tuning the model by updating the frozen reference model $\llm_{\text{ref}}$ with the fine-tuned model $\llm_{\theta}$ and re-estimating the data-generating process $\dgp^{(t)}$ and the expected-utility-maximizing decision $\decision^{*(t)}(\cdot, \cdot)$.
% We stop the iteration when the complementary value of the signals converges.
% We set the training epoch in each iteration to $2$ and stop the iteration when the improvement of the complementary value is less than $0.005$.

\mvspace{-2mm}
\section{Experiments}
\mvspace{-2mm}

We evaluate the value of \sysname{} in assisting decision-making in three steps.
First, we test whether \sysname{} can \textbf{recover complementary signals} when such signals are known by construction.
Second, we evaluate the \textbf{theoretical complementary value} of the extracted signals in three real-world decision scenarios: medical diagnosis, content moderation, and scientific paper reviewing.
Third, we test the \textbf{practical value} of the signals through qualitative interviews with domain experts, a decision study with human participants, and an LLM decision study.
Across all experiments, we instantiate the ``two agents'' setting with the supervisor's information $\supervisorinfo$, recommending agent's features $\agentfeature$, and payoff state $\payoffstate$.

\mvspace{-2mm}
\subsection{Model and Training Details}
\mvspace{-2mm}

% We use \textit{Qwen3-8B} to generate the signals for estimating the data-generating process and as the backbone language model for fine-tuning.
% For SFT, we train with 2 epochs, with a learning rate of $5 \times 10^{-6}$ and a cosine learning rate scheduler.
% For GRPO, we initialize the training with 10 epochs but stop early when the improvement of the reward on the validation set is less than $0.01$, with a learning rate of $1 \times 10^{-6}$ and a cosine learning rate scheduler.
% We sample 12 candidate completions for each instance in GRPO.
% For the hyperparameters that determine the threshold of the sample size $N_{\tau}$ for the data-generating process estimation, we use $\epsilon = 0.1$ and $\delta = 0.05$.
% We use the output token length of 1,500 tokens.

We use \textit{Qwen3-8B} as both the signal-generation reference model and the backbone for fine-tuning. 
We train with SFT followed by GRPO.
Implementation details, hyperparameters, and prompts are provided in \Cref{app:hyper,app:prompts}.

\mvspace{-2mm}
\subsection{Baselines \& Benchmark}
\mvspace{-2mm}

We compare \sysname{} against four interpretable baselines: zero-shot prompting, few-shot prompting, BERTopic as a concept-generation baseline (\textit{agnostic to the existing agent decisions and the decision problems}), and HypotheSAEs as a hypothesis-generation baseline (\textit{agnostic to the existing agent decisions}). 
We also report a non-interpretable benchmark obtained by fine-tuning Qwen3-8B to predict the state directly from $t$ and $z$. 
All methods use the same train/validation/test splits.
Additional baseline details are in \Cref{app:hyper_baseline}.

\mvspace{-2mm}
\subsection{Tasks, Datasets, and Metrics}
% \mvspace{-2mm}

\subsubsection{Recovering Complementary Signals by Construction}
% \mvspace{-2mm}

We first test how well \sysname{} recovers artificially induced complementary signals using the medical decision problem of diagnosing cardiac dysfunction (\Cref{tab:example_notation}).

\mvspace{-2mm}
\paragraph{Dataset and task.}
We use radiology reports as $\supervisorinfo$ from the MIMIC-CXR dataset~\citep{johnson2019mimic,johnson2023mimic,Johnson2024mimiccxr}, with a ground-truth signal set $\latentsignal^*$ derived from CheXpert labels~\citep{irvin2019chexpert} on these reports.
To generate the state $\payoffstate$ and upstream recommendation $\agentdecision$ for each instance, we fit a logistic regression model that regresses real-world cardiac dysfunction labels on the CheXpert labels.
We derive the cardiac dysfunction labels from blood tests related to cardiac dysfunction, specifically \textit{troponin} and \textit{NT-proBNP}, in MIMIC-IV, using domain-suggested age cutoffs~\citep{mueller2019heart, heidenreich20222022} to threshold them into binary values.

We generate $\payoffstate$ by inputting the full set of CheXpert report labels into this model.
We generate $\agentdecision$ by holding out a subset of labels, \textit{Edema} and \textit{Pleural Effusion}, so that they are predictive of $\payoffstate$ but not represented in $\agentdecision$ by construction.
These held-out labels therefore serve as ground-truth complementary signals.
To match the realistic setting where doctors are assisted by a probabilistic prediction, we threshold the model prediction at $0.5$ to obtain a binary state $\payoffstate \in \{0,1\}$, while keeping $\agentdecision \in [0,1]$ as the probabilistic recommendation.
We use 6K/2K/4K instances for training, validation, and testing.

\mvspace{-2mm}
\paragraph{Metrics.}
We evaluate both whether the extracted signals recover those signals (following prior work~\cite{zhong2024explaining}) and whether they improve decision-making beyond the upstream recommendation.

\begin{itemize}[leftmargin=*]
    \mvspace{-2mm}
    \item \textbf{Surface Similarity}: For each test instance, we prompt Qwen3-14B to score the surface similarity between each presented ground-truth signal, i.e., a ground-truth signal with label 1 for that instance, and each output signal.
    Specifically, Qwen3-14B assigns scores of 1, 0.5, and 0 for signals that are the same, related, or distinct, respectively.
    For stability, we report the average similarity score over 7 repetitions with temperature 0.7.
    For each presented ground-truth signal, we take the maximum similarity over output signals, corresponding to the best match, and average these per-ground-truth maxima across all instances.

    % \mvspace{-1mm}
    \item \textbf{F1 Similarity}: For each instance, we compute the F1 score between the presence of the ground-truth signals and whether an output signal is matched to each ground-truth signal, defined as having surface similarity $\geq 0.5$.
    We report the average F1 score over all instances and all ground truths.

    % \mvspace{-1mm}
    \item \textbf{Complementary Information Value}: We report the complementary information value $\rational^{\scoringrule, \agentdecisionRV}(\llm)$ from \Cref{eq:rational}, using accuracy $\scoringrule(\decision, \payoffstate) = \mathds{1}\left[\decision = \payoffstate\right]$ as the decision payoff.\footnote{We threshold the probabilistic decision at 0.5 to obtain a binary decision.}
    This measures the improvement in best-attainable performance from observing the extracted signals $\llm(\supervisorinfoRV)$ in addition to the upstream recommendation $\agentdecisionRV$.
    We use a multivariate logit model to fit the payoff state $\payoffstate$ on the extracted signals $\llm(\supervisorinfoRV)$ and the upstream recommendation $\agentdecisionRV$, and use this model to estimate the best-attainable performance.
\end{itemize}

% The first two metrics follow \citep{zhong2024explaining}, who similarly evaluate recovery of ground-truth signals.
% The third metric evaluates whether the recovered signals provide complementary information for the downstream decision problem.
\mvspace{-2mm}
\subsubsection{Theoretical Complementary Value in Real-World Decision Tasks}
% \mvspace{-2mm}

We next evaluate whether \sysname{} extracts signals with theoretical complementary value in three decision-making tasks and datasets: medical diagnosis, content moderation, and scientific paper reviewing.
% Unlike the construction-based setting above, these real-world tasks do not provide ground-truth complementary signals.
% We therefore evaluate the extracted signals by measuring whether they improve the best-attainable decision performance beyond the upstream recommendation and by measuring how many extracted signals add unique information beyond that recommendation.
Additional dataset details are provided in \Cref{app:data}.

\mvspace{-2mm}
\paragraph{Medical Diagnosis: MIMIC-CXR~\citep{Johnson2024mimiccxr}.}
We investigate what information in human-generated radiology reports can improve decisions over a vision model's predictions of cardiac dysfunction from chest X-ray images.
We use the radiology reports from MIMIC-CXR as $\supervisorinfo$, the chest X-ray images from MIMIC-CXR as $\agentfeature$, and the cardiac dysfunction labels derived from MIMIC-IV blood tests as $\payoffstate \in \{0,1\}$, following the construction described above.
We fine-tune the CXR foundation model~\citep{sellergren2023generalized} to predict cardiac dysfunction from chest X-ray images and blood-test-derived labels, and use its probabilistic predictions as the upstream recommendation $\agentdecision \in [0,1]$.
We use the same training, validation, and test split as in the construction-based experiment.

\mvspace{-2mm}
\paragraph{Content Moderation: DICES~\citep{aroyo2023dices}.}
We investigate what toxicity cues in human-LLM conversations are missed by a specific group of human annotators relative to the majority-vote average annotation.
We use the DICES dataset, which contains 115K human toxicity annotations on 1.3K human-LLM conversations, along with annotator demographic information.
We use annotations from the largest demographic group of annotators, Asian millennial women with a college degree or higher, as the upstream recommendation $\agentdecision \in \{0,1\}$ for non-toxic and toxic, respectively.
We use the majority-vote annotation from the entire population as the state $\payoffstate \in \{0,1\}$, and use the conversation text as $\supervisorinfo$.
We use 0.8K/0.2K/0.3K conversations for training, validation, and testing.

\mvspace{-2mm}
\paragraph{Scientific Paper Reviewing: Review5K~\citep{weng2025cycleresearcher}.}
We investigate what information in human-written reviews is missed by an LLM-generated review decision.
We use the Review5K dataset, which contains 5K scientific papers with human-written reviews and final accept/reject decisions.
We generate the LLM review by prompting Gemini 2.0 Flash to review each paper and provide a judgment among ``Accept'', ``Unsure'', and ``Reject''.
We use the human-written review text as $\supervisorinfo$, and the final judgment of the LLM review as the upstream recommendation $\agentdecision \in \{0, 0.5, 1\}$ for ``Reject'', ``Unsure'', and ``Accept'', respectively.
We use the final decision made by the human area chair in the real review process as the state $\payoffstate \in \{0,1\}$ for ``Reject'' and ``Accept'', respectively.
We use 3K/1K/1K instances for training, validation, and testing.

\mvspace{-2mm}
\paragraph{Metrics.}
We report two metrics that quantify the complementary value of the extracted signals.

\begin{itemize}[leftmargin=*]
    \mvspace{-2mm}
    \item \textbf{Complementary Information Value}: We use the same complementary information value metric as in the construction-based experiment.
    % This measures the improvement in best-attainable decision performance from observing the extracted signals in addition to the upstream recommendation.

    \mvspace{-1mm}
    \item \textbf{Breadth}: We fit a multivariate logit model of $\payoffstate$ on $\agentdecision$ and $\llm(\supervisorinfo)$, and report the number of extracted signals whose coefficients are significantly non-zero\footnote{We use a Bonferroni-corrected p-value threshold of $5\times 10^{-2}$.} when controlling for $\agentdecision$.
    % This measures how many extracted signals add unique information beyond the upstream recommendation.
\end{itemize}

\mvspace{-2mm}
\subsubsection{Practical Value in Downstream Decision-Making}

% Finally, we evaluate whether the complementary signals generated by \sysname{} are useful in practice.
% We do so through three evaluations: qualitative interviews with domain experts in the medical diagnosis task, a controlled human decision study in the content moderation task, and a prompting-based LLM experiment in the scientific paper reviewing task.

% \mvspace{-2mm}
\paragraph{Qualitative expert interviews.}
To assess whether the extracted medical signals are meaningful to domain experts, we conducted interviews with two practicing physicians: one cardiologist (P1) and one internist (P2), both of whom are also professors of medicine.
During each session, we walked through four patient cases in depth.
For each case, we first showed the radiology report, chest X-ray image, and vision model prediction, and asked the physician whether they agreed with the model prediction and what evidence from the report or image informed their judgment.
We then revealed the complementary signals generated by \sysname{} and asked the physician to assess their relevance and alignment with their domain knowledge.
Finally, we presented an updated prediction from a multivariate logit model that predicts cardiac dysfunction from the surfaced signals and the upstream agent decision, and asked whether the physician trusted this updated prediction more or less than the original prediction.

\begin{figure*}[t]
    \centering
    \includegraphics[width=\textwidth]{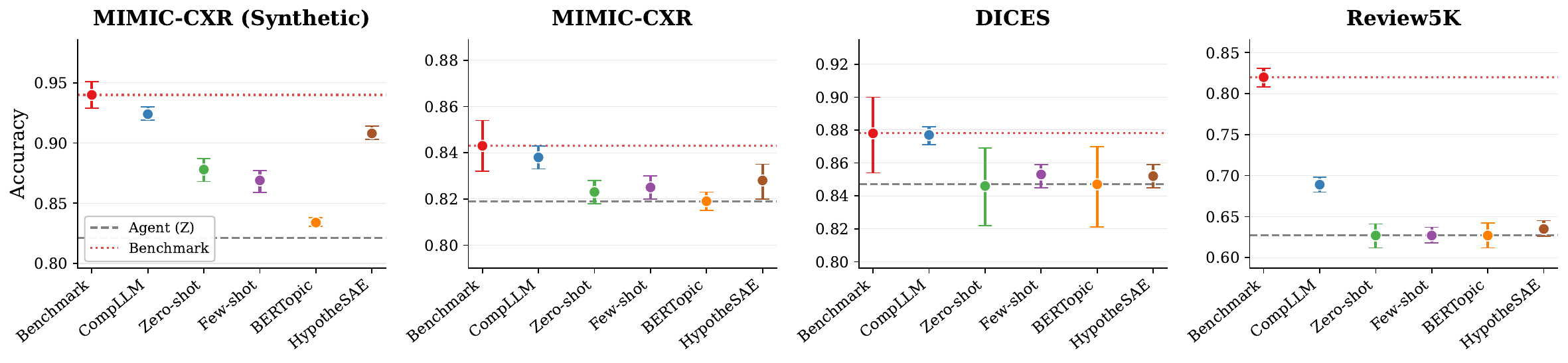}
    \mvspace{-6mm}
    \caption{Theoretical best-attainable accuracy given the extracted signals and the agent's recommendation by each method. The grey dashed lines represent agent decision accuracy. Error bars depict bootstrapped 95\% confidence intervals (N=5000).}
    \mvspace{-4mm}
    \label{fig:exp_result}
\end{figure*}

\mvspace{-2mm}
\paragraph{Human decision study.}
We conducted a within-subject human study to evaluate whether complementary signals generated by \sysname{} improve downstream decision-making in content moderation.
We recruited 232 participants from Prolific and asked each participant to make decisions on $28$ human-LLM conversations from the DICES dataset.
The 28 conversations were divided into four sets of seven conversations, each of which was randomly assigned to one of four decision-assistance conditions: no additional signal, where participants only saw the conversation and the upstream recommendation (i.e., annotations from Asian millennial women with a college degree or
higher); signals from \textsc{HypotheSAEs}; signals from \sysname{}; and direct predictions from the non-interpretable benchmark model used in \Cref{fig:exp_result}.
We estimate decision accuracy for each assistance condition according to our preregistered analysis plan, using a Bayesian hierarchical model.\footnote{The model is preregistered at \url{https://aspredicted.org/db3pm5.pdf}.}
See more details in \Cref{app:human_study}.

\mvspace{-2mm}
\paragraph{LLM decision study.}
We also evaluated whether \sysname{} signals improve LLM decision-making in the Review5K scientific paper reviewing task.
For each instance, we prompted an LLM to judge whether the paper would be accepted under three information conditions: the original paper only; the original paper with human-written reviews; and the original paper with human-written reviews plus complementary signals generated by \sysname{}.
We report the accuracy of the LLM judgments under each condition.

\mvspace{-2mm}
\section{Results}
\mvspace{-2mm}

\subsection{Recovering Complementary Signals by Construction}
\mvspace{-2mm}

\begin{wraptable}{r}{5cm}
    \centering
    \mvspace{-5mm}
    \caption{Results of average surface similarity and F1 score on the synthetic dataset.}
    \label{tab:synthetic_exp_results}
    % \resizebox{0.3\linewidth}{!}{
        \begin{tabular}{lcc}
            \toprule
            
            \textbf{Method} 
            & \textsc{Surf.} & \textsc{F1} \\
            \midrule

            \sysname{}
            & \textbf{0.98} & \textbf{0.67} \\
            
            \textsc{Zero-shot}
            & 0.91 & 0.42 \\
            
            \textsc{Few-shot}
            & 0.95 & 0.38 \\
            
            \textsc{BERTopic}
            & 0.84 & 0.17 \\
            
            \textsc{HypotheSAE}
            & 0.90 & 0.38 \\
            
            \bottomrule
            \end{tabular}
    % }
    % \mvspace{-8mm}
\end{wraptable}

\sysname{} best recovers the construction-based complementary signals.
It achieves the highest surface similarity and F1 similarity among all methods (\Cref{tab:synthetic_exp_results}), and its extracted signals provide the largest complementary information value among interpretable methods (\Cref{fig:exp_result}).
% As shown in \Cref{tab:synthetic_exp_results}, \textbf{\sysname{} outperforms all baseline methods in both surface similarity and F1 similarity}.
The qualitative examples in \Cref{tab:synthetic_signals} further show that the signals extracted by \sysname{} \textbf{cover the ground-truth complementary signals}.
We also find that \textbf{fine-tuning is important for recovering instance-level complementary signals}.
Although few-shot prompting produces signals that are more semantically similar to the ground-truth signals than zero-shot prompting, it does not reliably identify the correct signals for the correct instances, as reflected by its low F1 score.
This suggests that prompt-only methods may recover plausible signal types but fail to localize them at the instance level.
% Finally, the signals extracted by \sysname{} provide the \textbf{highest complementary information value} over the upstream agent's recommendation among all interpretable methods in the construction-based dataset, as shown in \Cref{fig:exp_result}.
% The only method with higher or comparable value is the non-interpretable benchmark, which directly optimizes predictive performance without producing human-interpretable signals.

% It achieves the highest surface similarity and F1 similarity among all methods (Table 2), and its extracted signals provide the largest complementary information value among interpretable methods (Figure 3). The gap between few-shot surface similarity and F1 suggests that prompting can recover plausible signal types but is less reliable at localizing the correct signals for each instance.

% \mvspace{-2mm}
\subsection{Theoretical Complementary Value in Real-World Decision Tasks}
% \mvspace{-2mm}

% \Cref{fig:exp_result} shows the complementary information value of signals extracted by each method across these datasets.
% \Cref{tab:signal_results_cxr,tab:signal_results_review5k,tab:signal_results_dices} report the significant signals identified in MIMIC-CXR, Review5K, and DICES, respectively.

\textbf{\sysname{} extracts signals with the highest complementary information value among interpretable methods.}
Compared with the baseline methods, \sysname{} is the only interpretable method that extracts signals with significant complementary information value in all three real-world datasets, as indicated by confidence intervals that do not overlap with the accuracy of the recommending agent in \Cref{fig:exp_result}.
This result suggests that explicitly optimizing for complementarity is necessary for improving over the existing agent's recommendation.
In \textsc{MIMIC-CXR} and \textsc{DICES}, \sysname{} achieves performance comparable to the non-interpretable benchmark, suggesting that extracting discrete, interpretable complementary signals does not substantially sacrifice predictive value in these two tasks.
At the same time, \sysname{} produces signals that remain inspectable and usable by downstream decision-makers, unlike the benchmark model.
\textbf{\sysname{} identifies a broader set of complementary signals.}
Across the three real-world datasets, \sysname{} extracts more signals with significant complementary information value than the baseline methods, as shown in \Cref{tab:signal_results_cxr,tab:signal_results_review5k,tab:signal_results_dices}.
% This indicates that the signals extracted by \sysname{} are more diverse and comprehensive, capturing multiple sources of information that add value beyond the upstream recommendation.

% \mvspace{-2mm}
\subsection{Practical Value in Downstream Decision-Making}
% \mvspace{-2mm}

\paragraph{Qualitative Feedback on \sysname{} for Medical Diagnosis}
The physicians viewed multiple overlapping cases (3 and 4 respectively, for a total of 6). Both \textbf{appraised the complementary signals as aligning with their domain knowledge}, with two exceptions. The first was a case where P1 noticed a signal that they believed provided relevant complementary information (the patient's history  of cardiac problems), but was missed by \sysname{}. 
Additionally, both physicians indicated that one of the complementary signals for a case (\Cref{fig:case1_info}), named \textit{negative\_pleural\_effusion} and referring to an absence of pleural effusion and pneumothorax (\Cref{fig:case1_sig}), was only half relevant, as the second condition was neither associated strongly with cardiac dysfunction or the most likely alternative condition.
In one other case, P2 did not dispute the complementary signals, but felt they might be consistent with--rather than adding new information beyond--the original prediction. %\ken{I'm wondering how to interpret the previous sentence. Was this a case where the complementary signal was directionally consistent with the original prediction? (i.e., probability was high and the signal should only increase it) If so, it is not clear that P2's observation should be noted as an 'exception' per se.}  
In an exit interview, both physicians were enthusiastic about the potential value for \sysname{} in practice, including to assist in creating lists of supporting versus contradictory evidence that clinicians already make (P1, P2), and to support ER doctors and general internists who have less specific training in cardiology (P1).   

% \mvspace{-2mm}
\paragraph{Improved Human Decisions on Content Moderation}
% \mvspace{-2mm}

\begin{figure}[t]
\centering
\begin{subfigure}[b]{0.49\textwidth}
    \centering
    \includegraphics[width=\textwidth]{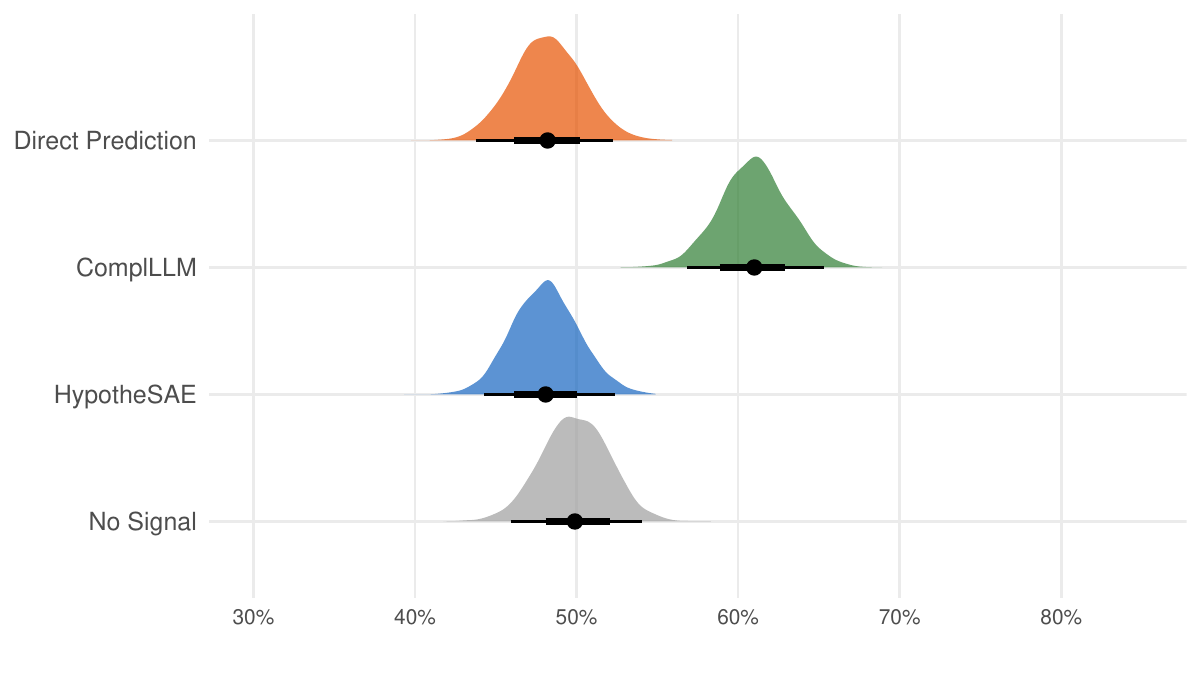}
    \mvspace{-5mm}
    \caption{Accuracy of human toxicity decisions for the DICES dataset, represented as posterior predictive distributions from our fitted model. Intervals depict 95\% and 75\% highest density intervals (HDIs).}
    \label{fig:human_study_content_moderation}
    % \mvspace{-8mm}
\end{subfigure}
\hfill
\begin{subfigure}[b]{0.49\textwidth}
    \centering
    \includegraphics[width=\textwidth]{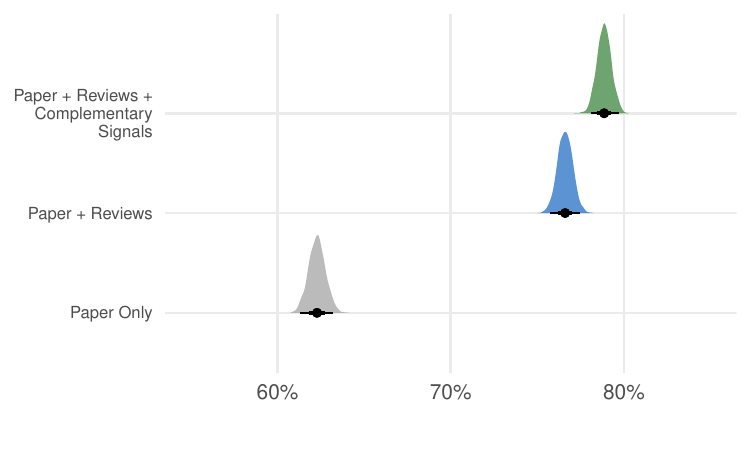}
    \mvspace{-5mm}
    \caption{Accuracy of LLM paper review decisions on the \textsc{Review5K} dataset, represented by the density curves of the bootstrapped samples (N=5000). Intervals depict 95\% and 75\% highest density intervals (HDIs).}
    % \mvspace{-4mm}
    \label{fig:paper_review_llm_performance}
\end{subfigure}
\caption{Practical value of signals generated by \sysname{} in downstream decision-making.}
\mvspace{-3mm}
\end{figure}

\Cref{fig:human_study_content_moderation} shows the accuracy of human annotators under different decision-assistance conditions for the content moderation task.
\sysname{} shows a clear advantage over other forms of decision support. Decisions made with assistance from \sysname{} signals led to an accuracy of 60.9\%, with 95\% HDI [56.7\%, 65.1\%], while the accuracy of all other conditions hovered near 50\%. 
Signals from \textsc{HypotheSAEs}, which achieved 48.0\% accuracy with 95\% HDI [44.0\%, 51.9\%], did not improve human decisions relative to the no-signal condition, which achieved 49.8\% accuracy with 95\% HDI [46.0\%, 53.9\%].
More importantly, direct predictions from the benchmark model did \textit{not} help participants achieve higher accuracy despite offering the most decision-relevant information for the task.
Participants in the direct prediction condition achieved 48.2\% accuracy, with 95\% HDI [43.8\%, 52.3\%].
This suggests that, despite providing the highest theoretical information value (\Cref{fig:exp_result}), human-interpretable complementary signals may be necessary for that information to translate to improved human decision-making.

% \mvspace{-2mm}
\paragraph{Improved LLM Decisions on Scientific Paper Reviewing}
% \mvspace{-2mm}

% We further evaluate the practical usefulness of complementary signals by providing them to a proxy LLM decision-maker, Gemini 2.0 Flash, in the scientific paper reviewing task.
\Cref{fig:paper_review_llm_performance} shows the accuracy of the LLM's accept/reject decisions relative to the human area chair's final decision in Review5K.
We find that adding \sysname{} signals improves LLM decision accuracy over providing the paper and human reviews alone.
The LLM with complementary signals achieves $78.9\%$ accuracy ([95\% HDI: $78.1\%$, $79.7\%$]), compared with $76.6\%$ accuracy ([95\% HDI: $75.8\%$, $77.5\%$]) when given only the paper and human-written reviews.

\mvspace{-2mm}
\section{Discussion and Limitations}
\label{sec:discussion}
\mvspace{-2mm}

AI models intended for decision support should present signals that are complementary to those a decision-maker already has access to. 
We propose a method to post-train LLMs to give complementary signals based on decision theory. 
Throughout three real-world decision tasks, we show that explicitly surfacing the complementary signals produced by our method offers meaningful cues to domain experts and improves the decision quality of with crowdsourced participants and LLM agents.

Complementary signals can help downstream agents use evidence already available to them.
Although an ideal Bayesian decision-maker with knowledge of the data-generating distribution will not benefit more from complementary signals once they already have access to the information $\supervisorinfoRV$ and the existing recommendation $\agentdecisionRV$, boundedly rational agents like humans and LLMs may struggle to identify which findings matter conditional on an upstream recommendation.
In our \textsc{Review5K} study, the downstream LLM already receives the paper and human reviews, but adding \sysname{} signals derived from the human reviews still improves its decisions.
% This result supports the usefulness of highlighting existing evidence, but does not establish that replacing the original inputs with a compressed representation would improve performance.

One interpretation is that \sysname{} makes task-specific statistical knowledge usable by other agents.
Its training objective targets signals with additional decision-relevant value conditional on the existing recommendation, which learns relationships that a downstream agent may not infer from an individual instance.
Our method expresses these findings in language to communicate this learned knowledge across heterogeneous agents.
Our human study illustrates why this interpretable form of assistance matters.
Direct predictions from a strong benchmark model did not improve the participants' accuracy, while complementary signals produced by \sysname{} did.

We also hypothesize that the benefit may depend on how well the downstream agent already knows the data-generating distribution of the decision task.
An agent trained directly on the text, upstream recommendation, and outcomes may have already internalized the relationships that our complementary signals serve to surface.
All of these possibilities are open for future work to evaluate.

\paragraph{Limitations.}
Signal discovery and presence labeling depend on the reference LLM and can inherit its omissions or labeling errors.
Frequency filtering can also discard rare findings that matter to domain experts.
The analyses in \Cref{app:robustness} assess the sensitivity of these dependencies.

Complementarity also depends on the recommending agent and task distribution, both of which may change.
When the recommending agent and supervisor are the same, viewing complementary signals may induce learning that changes subsequent judgments and the value of future assistance.
Periodically re-estimating these relationships and updating the extractor could address such changes, but requires evaluation as a continual learning problem.
% \footnote{The data and code to reproduce our experimental results are available at \url{https://osf.io/z5gyj/overview?view_only=c32fcf0a14b54b9fa6ed22ddb8d1f774}.}

% \section*{Code and Data}

\bibliography{ref}
\bibliographystyle{plain}

%%%%%%%%%%%%%%%%%%%%%%%%%%%%%%%%%%%%%%%%%%%%%%%%%%%%%%%%%%%%%%%%%%%%%%%%%%%%%%%
%%%%%%%%%%%%%%%%%%%%%%%%%%%%%%%%%%%%%%%%%%%%%%%%%%%%%%%%%%%%%%%%%%%%%%%%%%%%%%%
% APPENDIX
%%%%%%%%%%%%%%%%%%%%%%%%%%%%%%%%%%%%%%%%%%%%%%%%%%%%%%%%%%%%%%%%%%%%%%%%%%%%%%%
%%%%%%%%%%%%%%%%%%%%%%%%%%%%%%%%%%%%%%%%%%%%%%%%%%%%%%%%%%%%%%%%%%%%%%%%%%%%%%%
\newpage
\appendix
\onecolumn

\begin{table}[!h]
    \centering
    \caption{Summary of Extracted Signals by Method on Synthetic Dataset.}
    \label{tab:synthetic_signals}
    \small
    \begin{tabular}{lcp{12cm}}
    \toprule
    \textbf{Method} & \textbf{Count} & \textbf{Signals} \\
    \midrule
    \multirow{2}{*}{Ground Truth} & \multirow{2}{*}{2} & 
    Edema \newline
    Pleural Effusion \\
    \addlinespace
    \multirow{10}{*}{\sysname{}} & \multirow{10}{*}{16} & 
    positive\_pleural\_effusion \newline
    positive\_edema \newline
    positive\_cardiomegaly \newline
    positive\_pleural\_effusion\_left \newline
    positive\_pleural\_effusion\_right \newline
    positive\_pneumonia \newline
    positive\_edema\_resolution \newline
    positive\_troponins \newline
    positive\_atelectasis \newline
    no\_pulmonary\_edema ... \\
    \addlinespace
    \multirow{8}{*}{\textsc{HypotheSAE}} & \multirow{8}{*}{8} & 
    mentions presence of pulmonary edema \newline
    mentions presence of interstitial edema \newline
    mentions presence of interstitial lung disease or interstitial findings \newline
    mentions no acute cardiopulmonary process \newline
    mentions presence of pleural effusion and atelectasis \newline
    mentions presence of pleural effusion \newline
    mentions presence of bilateral pleural effusion \newline
    mentions presence of interstitial pulmonary edema \\
    \addlinespace
    \multirow{10}{*}{\textsc{BERTopic}} & \multirow{10}{*}{15} & 
    Outlier/Noise \newline
    Chest X-ray Findings \newline
    Cardiac Enlargement \newline
    Pancreatic Cancer and Chest Imaging \newline
    Chest X-ray for TIA evaluation \newline
    Chest X-ray for Upper GI Bleed \newline
    Pulmonary Edema Monitoring \newline
    Pulmonary Edema Evaluation \newline
    Chest X-ray findings in edema \newline
    Chest X-ray findings ... \\
    \addlinespace
    \multirow{10}{*}{\textsc{Few-shot}} & \multirow{10}{*}{26} & 
    enlarged\_cardiac\_silhouette \newline
    pulmonary\_vascular\_congestion \newline
    pleural\_effusion \newline
    vascular\_congestion \newline
    moderate\_cardiomegaly \newline
    bilateral\_pleural\_effusions \newline
    interstitial\_edema \newline
    cardiomegaly \newline
    mild\_pulmonary\_edema \newline
    pulmonary\_congestion ... \\
    \addlinespace
    \multirow{10}{*}{\textsc{Zero-shot}} & \multirow{10}{*}{55} & 
    pulmonary\_vascular\_congestion \newline
    cardiac\_enlargement \newline
    pleural\_effusions \newline
    cardiac\_silhouette\_enlarged \newline
    bilateral\_pleural\_effusions \newline
    small\_left\_pleural\_effusion \newline
    no\_cardiomegaly \newline
    no\_pleural\_effusion \newline
    moderate\_cardiomegaly \newline
    pulmonary\_edema ... \\
    \bottomrule
    \end{tabular}
    \end{table}

    \begin{table*}[!h]
        \centering
        \caption{Signals on MIMIC-CXR dataset that are significant with $p < 0.05$ after Bonferroni correction. `$\Delta$ Acc.' represents the marginal accuracy improvement each signal $\latentsignal_j$ marginally provides over the other signals and the agent recommendation. `Total Acc.' shows the combined accuracy when all significant signals are included, i.e., sum of the signals' marginal gain. `\# Sig' indicates the number of statistically significant signals discovered by each method.}
        % \mvspace{-2mm}
        \label{tab:signal_results_cxr}
        \resizebox{\linewidth}{!}{
        \begin{tabular}{lcclc}
        \toprule
        \textbf{Source} & \textbf{Total Acc.} & \textbf{\# Sig} & \textbf{Significant Signals} & \textbf{$\Delta$ Acc.} \\
        \midrule
        Agent Decision & 0.819 & \textit{-} & \textit{-} & \textit{-} \\
        \midrule
        \multirow{12}{*}{\textsc{ComplLLM}} & \multirow{12}{*}{0.839} & \multirow{12}{*}{12} & \textit{negative pneumothorax} & +0.0048 \\
         & & & \textit{positive pleural effusion} & +0.0040 \\
         & & & \textit{negative pleural effusion} & +0.0032 \\
         & & & \textit{positive pulmonary congestion} & +0.0024 \\
         & & & \textit{positive cardiac silhouette enlargement} & +0.0016 \\
         & & & \textit{positive cardiac enlargement} & +0.0008 \\
         & & & \textit{uncertain pleural effusion} & +0.0008 \\
         & & & \textit{uncertain cardiomegaly} & +0.0006 \\
         & & & \textit{positive pulmonary vascular congestion} & +0.0006 \\
         & & & \textit{negative pulmonary edema} & +0.0002 \\
         & & & \textit{negative cardiac enlargement} & +0.0002 \\
         & & & \textit{positive cardiomegaly} & +0.0002 \\
        \midrule
        \multirow{6}{*}{\textsc{HypotheSAE}} & \multirow{6}{*}{0.831} & \multirow{6}{*}{6} & \textit{mentions no pleural effusion} & +0.0032 \\
         & & & \textit{mentions absence of pleural effusion or pneumothorax} & +0.0024 \\
         & & & \textit{mentions pulmonary edema} & +0.0020 \\
         & & & \textit{mentions presence of pleural effusion} & +0.0020 \\
         & & & \textit{mentions presence of pulmonary edema} & +0.0018 \\
         & & & \textit{mentions pulmonary edema or interstitial opacities} & +0.0008 \\
        \midrule
        \multirow{5}{*}{\textsc{Zero-Shot}} & \multirow{5}{*}{0.825} & \multirow{5}{*}{5} & \textit{cardiac enlargement} & +0.0014 \\
         & & & \textit{pleural effusion} & +0.0012 \\
         & & & \textit{bilateral pleural effusion} & +0.0012 \\
         & & & \textit{interstitial edema} & +0.0010 \\
         & & & \textit{pulmonary vascular engorgement} & +0.0010 \\
        \midrule
        \multirow{3}{*}{\textsc{Few-Shot}} & \multirow{3}{*}{0.823} & \multirow{3}{*}{3} & \textit{no pleural effusion} & +0.0014 \\
         & & & \textit{positive interstitial edema} & +0.0012 \\
         & & & \textit{positive cardiac silhouette enlargement} & +0.0010 \\
        \midrule
        \textsc{BERTopic} & 0.818 & 1 & \textit{endotracheal tube position} & +0.0005 \\
        \bottomrule
        \end{tabular}
        }
        \end{table*}

\begin{table*}[!h]
    \centering
    \caption{Signals on DICES dataset. Significant signals with $p < 0.05$ after Bonferroni correction. `$\Delta$ Acc.' represents the incremental accuracy improvement each signal provides over the baseline agent decision (0.847). `Total Acc.' shows the combined accuracy when all significant signals are included. `\# Sig' indicates the number of statistically significant signals discovered by each method.}
    \label{tab:signal_results_dices}
    \vspace{0.5em}
    \resizebox{\linewidth}{!}{
    \begin{tabular}{lcclc}
    \toprule
    \textbf{Source} & \textbf{Total Acc.} & \textbf{\# Sig} & \textbf{Significant Signals} & \textbf{$\Delta$ Acc.} \\
    \midrule
    \multicolumn{5}{c}{\textit{Baseline Agent Decision Accuracy: 0.847}} \\
    \midrule
    \multirow{4}{*}{\textsc{ComplLLM}} & \multirow{4}{*}{0.878} & \multirow{4}{*}{4} & \textit{promotes or condones violence} & +0.0222 \\
     & & & \textit{misinformation conspiracy or false theory} & +0.0061 \\
     & & & \textit{offers safe alternative} & +0.0020 \\
     & & & \textit{oblique or profane language} & +0.0010 \\
    \midrule
    \multirow{2}{*}{\textsc{BERTopic}} & \multirow{2}{*}{0.856} & \multirow{2}{*}{2} & \textit{Outlier/Noise} & +0.0051 \\
     & & & \textit{Chat conversations} & +0.0030 \\
    \midrule
    \multirow{2}{*}{\textsc{HypotheSAE}} & \multirow{2}{*}{0.854} & \multirow{2}{*}{2} & \textit{contains explicit insults or profanity} & +0.0051 \\
     & & & \textit{mentions `he' or `him' in context of another person's actions} & +0.0010 \\
    \midrule
    \textsc{Zero-Shot} & 0.847 & 0 & \textit{-} & - \\
    \midrule
    \textsc{Few-Shot} & 0.847 & 0 & \textit{-} & - \\
    \bottomrule
    \end{tabular}
    }
    \end{table*}
    
    % Required packages in preamble:
    % \usepackage{booktabs}
    % \usepackage{multirow}

\begin{table*}[!h]
    \centering
    \caption{Signals on Review5k dataset. Significant signals with $p < 0.05$ after Bonferroni correction. `$\Delta$ Acc.' represents the incremental accuracy improvement each signal provides over the baseline agent decision (0.627). `Total Acc.' shows the combined accuracy when all significant signals are included. `\# Sig' indicates the number of statistically significant signals discovered by each method.}
    \label{tab:signal_results_review5k}
    \vspace{0.5em}
    \resizebox{\linewidth}{!}{
    \begin{tabular}{lcclc}
    \toprule
    \textbf{Source} & \textbf{Total Acc.} & \textbf{\# Sig} & \textbf{Significant Signals} & \textbf{$\Delta$ Acc.} \\
    \midrule
    Agent Decision & 0.627 & \textit{-} & \textit{-} & \textit{-} \\
    \midrule
    \multirow{8}{*}{\textsc{ComplLLM}} & \multirow{8}{*}{0.692} & \multirow{8}{*}{8} & \textit{clarity} & +0.0310 \\
     & & & \textit{novelty} & +0.0179 \\
     & & & \textit{experimental validation} & +0.0110 \\
     & & & \textit{theoretical contribution} & +0.0024 \\
     & & & \textit{clarity of presentation} & +0.0021 \\
     & & & \textit{insufficient comparison with related work} & +0.0002 \\
     & & & \textit{missing baselines} & +0.0002 \\
     & & & \textit{novelty limited} & +0.0002 \\
    \midrule
    \multirow{6}{*}{\textsc{HypotheSAE}} & \multirow{6}{*}{0.649} & \multirow{6}{*}{6} & \textit{mentions evaluations on specific datasets and comparisons} & +0.0058 \\
     & & & \textit{mentions graph-based techniques for efficiency or scalability} & +0.0053 \\
     & & & \textit{mentions analysis of computational complexity} & +0.0043 \\
     & & & \textit{mentions diffusion models} & +0.0041 \\
     & & & \textit{mentions theoretical analysis and experimental validation} & +0.0030 \\
     & & & \textit{mentions methodology compared to existing approaches} & +0.0010 \\
    \midrule
    \textsc{BERTopic} & 0.627 & 1 & \textit{Test-Time Adaptation} & +0.0002 \\
    \midrule
    \textsc{Zero-Shot} & 0.627 & 0 & \textit{-} & - \\
    \midrule
    \textsc{Few-Shot} & 0.627 & 0 & \textit{-} & - \\
    \bottomrule
    \end{tabular}
    }
    \end{table*}
    
    % Required packages in preamble:
    % \usepackage{booktabs}
    % \usepackage{multirow}

\newpage

\section{Human Study on Content Moderation}
\label{app:human_study}

To empirically evaluate whether \sysname{} signals improve human decision-making, we conducted a preregistered within-subject experiment in which participants predicted the toxicity of online conversations under four signal conditions. The study was preregistered at \url{https://aspredicted.org/db3pm5.pdf}.

\subsection{Task}

Participants were shown 28 conversations between humans and an LLM, and asked to predict whether the final response in each conversation was \textbf{Toxic} or \textbf{Not Toxic}. 
Each conversation was drawn from \textsc{DICES} dataset.
After each prediction, we also asked the participants to rate their confidence on a 1-5 Likert scale.

\subsection{Conditions}
Every participant experienced all four conditions in a counterbalanced order.
In all conditions, participants saw the conversation text and an existing annotation from the demographic group of annotators, Asian millennial women with a college degree or higher. 
The conditions differed in what additional AI-generated hint, if any, was provided to help participants catch labeling errors:

\begin{enumerate}
    \item \textbf{No Signal}: Participants made predictions without any AI assistance, relying solely on their own judgment.
    \item \textbf{Signals from \textsc{HypotheSAEs}}: Participants saw interpretable features extracted by an AI system identifying specific patterns in the conversation (e.g., presence of explicit insults, mentions of public figures, vaccine skepticism). These correspond to the signal-level features from our framework (agnostic to the existing decisions).
    \item \textbf{\sysname{} Signals}: Participants saw brief explanations generated by the complementary LLM identifying the most relevant behavioral pattern conditioned on the existing decisions in the final response (e.g., ``promotes violence,'' ``offers safe alternative'') along with a short excerpt and a one-sentence explanation of the pattern.
    \item \textbf{Direct AI Prediction}: Participants saw a binary Toxic/Not Toxic label predicted by an AI system, with no interpretive explanation. This condition serves as a baseline representing a standard AI-assisted setup where the human receives the model's prediction but no interpretable information.
\end{enumerate}

\subsection{Design}

The study used a fully within-subject design. Each participant completed four blocks of 7 trials each (28 trials total), one block per condition. The order of conditions was randomized across participants using one of the 24 possible permutations ($4! = 24$), ensuring full counterbalancing of order effects. 
Within each block, the 7 conversation instances were randomly sampled and shuffled from the dataset independently for each participant.

Each block began with an introduction page explaining whether and what type of signal would be provided. The signal (if any) was displayed alongside the conversation text on each trial page.
See the screenshots of the interfaces in \Cref{app:screenshots_human_study}.

\subsection{Participants}

We recruited 232 participants via Prolific. All participants were located in the United States and had not completed any pilot version of the experiment. Participants were informed that the study involved evaluating conversations that may contain profane, offensive, or sexually suggestive language, and provided informed consent before proceeding.

\subsection{Compensation}

Participants received a base payment of \$3.00 plus a \$0.10 bonus for each correct prediction out of 28, for a total possible payment ranging from \$3.00 to \$5.80. This incentive-compatible payment structure was designed to motivate careful engagement with the task.

\subsection{Exclusion Criteria}

Following our preregistration, we planned to exclude participants who: (a) did not complete all 28 predictions, (b) selected the same answer for all 7 instances within more than one block, or (c) had a median response time below 5 seconds per instance. % [Report how many were excluded, if any.]

\subsection{Statistical Model}

We fit a Bayesian mixed-effects logistic regression using the \texttt{brms} package \citep{burkner2017brms} in R:
\begin{equation}
\begin{aligned}
\texttt{correct} \sim & \texttt{signal\_condition} + \texttt{block\_position} +\\ & (1 + \texttt{signal\_condition} \mid \texttt{participant\_id}) + (1 \mid \texttt{instance\_id})
\end{aligned}
\label{eq:glmm}
\end{equation}
where:
\begin{itemize}
    \item \texttt{correct} is a binary outcome (1 = prediction matches ground truth, 0 = incorrect).
    \item \texttt{signal\_condition} is a four-level factor (reference: no signal) with levels: hypothesis, \textsc{ComplLLM}, and direct.
    \item \texttt{block\_position} (1--4) controls for order and fatigue effects.
    \item $(1 + \texttt{signal\_condition} \mid \texttt{participant\_id})$ allows each participant to have their own baseline accuracy and their own sensitivity to each signal type.
    \item $(1 \mid \texttt{instance\_id})$ accounts for varying difficulty across conversation instances.
\end{itemize}

We used weakly informative priors: $\mathcal{N}(0, 1.5)$ on fixed effects, Half-$\mathcal{N}(0, 1)$ on random effect standard deviations, and $\text{LKJ}(2)$ on the correlation matrix for random slopes. The model was fit with 4 chains of 4{,}000 iterations (1{,}000 warmup) using the NUTS sampler. All $\hat{R}$ values were at or below 1.01, and no divergent transitions were observed.

\subsection{Additional Results}

\subsubsection{Pairwise Contrasts}
 
Table~\ref{tab:pairwise} reports all pairwise contrasts between conditions.
 
\begin{table}[h]
\centering
\caption{Pairwise contrasts between conditions. Differences are on the log-odds scale. CrI = credible interval. $P(\text{diff} > 0)$ = posterior probability that the first condition has higher accuracy than the second.}
\label{tab:pairwise}
\begin{tabular}{lcccc}
\toprule
\textbf{Contrast} & \textbf{Diff} & \textbf{OR} & \textbf{95\% CrI} & $P(\text{diff} > 0)$ \\
\midrule
\textsc{ComplLLM} vs.\ No Signal              & $0.45$ & 1.57 & [1.34, 1.84] & $> 99.9\%$ \\
\textsc{ComplLLM} vs.\ \textsc{HypotheSAE}    & $0.53$ & 1.69 & [1.44, 1.99] & $> 99.9\%$ \\
\textsc{ComplLLM} vs.\ Direct Prediction       & $0.52$ & 1.68 & [1.44, 1.98] & $> 99.9\%$ \\
\textsc{HypotheSAE} vs.\ No Signal            & $-0.07$ & 0.93 & [0.81, 1.07] & $16.2\%$ \\
\textsc{HypotheSAE} vs.\ Direct Prediction    & $-0.01$ & 0.99 & [0.86, 1.15] & $47.3\%$ \\
No Signal vs.\ Direct Prediction               & $0.07$ & 1.07 & [0.93, 1.25] & $81.4\%$ \\
\bottomrule
\end{tabular}
\end{table}
 
All three pairwise contrasts involving \textsc{ComplLLM} showed posterior evidence ($> 99.9\%$). 
The contrast between \textsc{ComplLLM} and \textsc{HypotheSAE} is strong (OR $= 1.69$, 95\% CrI [1.44, 1.99]).
The remaining three contrasts (among \textsc{HypotheSAE}, No Signal, and Direct Prediction) showed no meaningful differences, indicating that these three conditions performed comparably.

\subsubsection{Order Effects}

Block position had a negligible effect on accuracy (coefficient $= -0.01$, 95\% CrI [$-0.13$, $0.12$]), indicating that the results are not confounded by fatigue or learning over the course of the experiment. The advantage of \textsc{ComplLLM} signals was stable across all four block positions.

\begin{figure}[h]
    \centering
    \includegraphics[width=\linewidth]{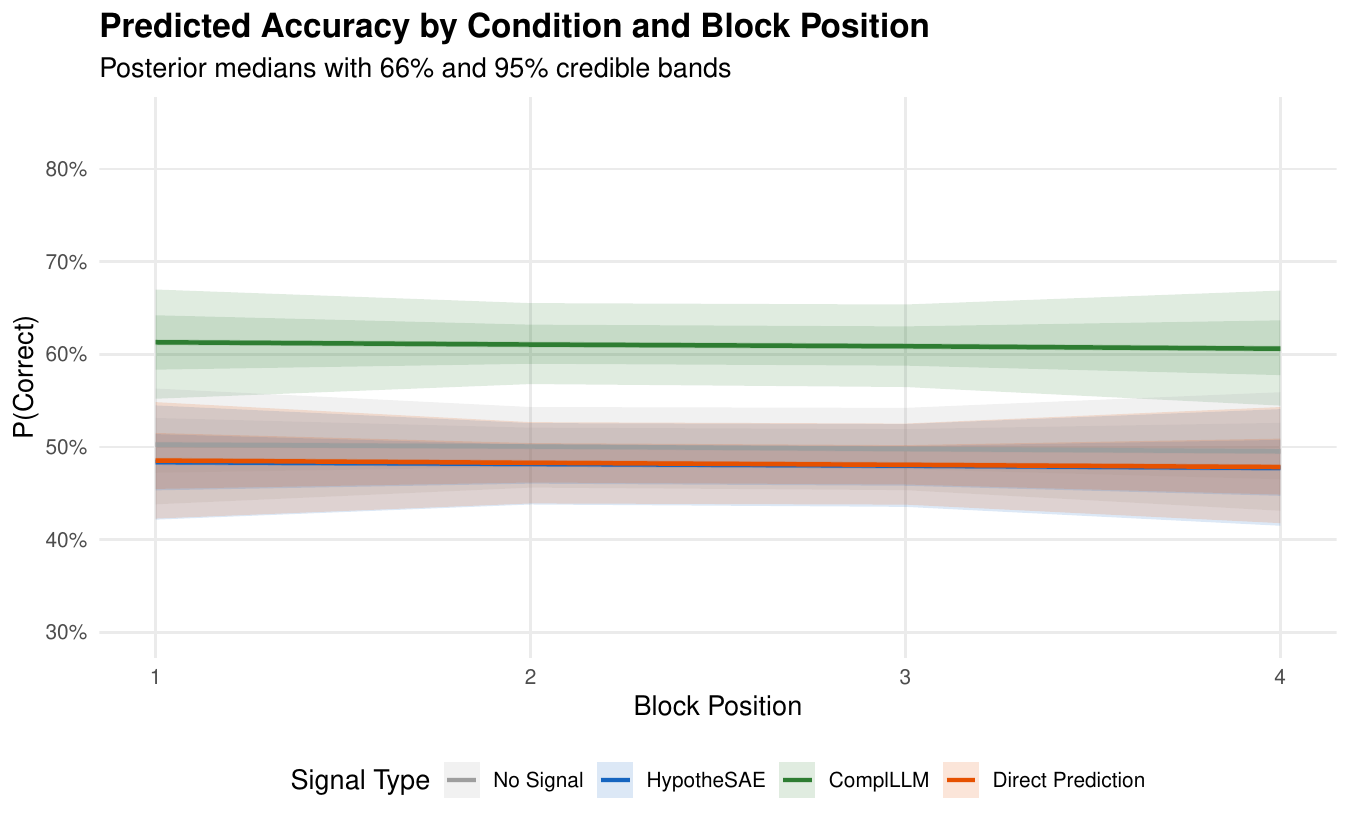}
    \caption{Predicted accuracy by condition and block position. Lines show posterior medians; shaded bands show 66\% and 95\% credible intervals. The advantage of \textsc{ComplLLM} signals is stable across all block positions.}
    \label{fig:accuracy_by_position}
\end{figure}

\subsubsection{Individual Differences}
 
Table~\ref{tab:random_effects} reports the random effect standard deviations from the Bayesian model, capturing individual variation across participants and instances.

\begin{table}[h]
\centering
\caption{Random effect standard deviations (log-odds scale) from the Bayesian mixed-effects model. Participant-level effects capture individual variation in baseline accuracy (Intercept) and sensitivity to each signal type (slopes). Instance-level effects capture variation in difficulty across conversations.}
\label{tab:random_effects}
\begin{tabular}{llcccc}
\toprule
\textbf{Group} & \textbf{Effect} & \textbf{SD} & \textbf{95\% CrI} \\
\midrule
\multirow{4}{*}{Participant} 
    & Intercept (baseline accuracy)           & 0.10 & [0.01, 0.24] \\
    & Slope: \textsc{HypotheSAE}              & 0.13 & [0.01, 0.34] \\
    & Slope: \textsc{ComplLLM}                & 0.46 & [0.21, 0.67] \\
    & Slope: Direct Prediction                & 0.28 & [0.03, 0.51] \\
\midrule
Instance & Intercept (instance difficulty)    & 0.48 & [0.38, 0.60] \\
\bottomrule
\end{tabular}
\end{table}
 
The participant-level intercept SD was small (0.10, 95\% CrI [0.01, 0.24]), indicating relatively homogeneous baseline accuracy across participants when no AI signal was provided. 
However, the random slope for \textsc{ComplLLM} was substantially larger (SD $= 0.46$, 95\% CrI [0.21, 0.67]) than the slopes for \textsc{HypotheSAE} (SD $= 0.13$, 95\% CrI [0.01, 0.34]) and Direct Prediction (SD $= 0.28$, 95\% CrI [0.03, 0.51]). 
This indicates that while \textsc{ComplLLM} signals improve accuracy on average, there is substantial individual variation in how much participants benefit, i.e., some participants improved dramatically with AI Explanations while others showed minimal gains.
 
The instance-level intercept SD was 0.48 (95\% CrI [0.38, 0.60]), reflecting meaningful variation in difficulty across conversations, i.e., some instances were consistently easier or harder to label correctly regardless of condition.

\section{Data Preprocessing}
\label{app:data}

\paragraph{\textsc{MIMIC-IV.}}
We use data from the MIMIC dataset~\citep{johnson2023mimic}, which contains anonymized electronic health records from Beth Israel Deaconess Medical Center (BIDMC), a large teaching hospital in Boston, Massachusetts affiliated with Harvard Medical School.
We download the X-ray images and corresponding reports of radiologists from the MIMIC-CXR dataset~\citep{johnson2019mimic}.
We join the MIMIC-CXR dataset with MIMIC-IV by matching on patient and visit ID, and filter the radiology images and reports to those for which there is at least one follow-up blood test related to cardiac dysfunction for the patient at the same visit.
We consider two types of blood tests following domain expert suggestions~\citep{mueller2019heart,heidenreich20222022}: \textit{Troponin} and \textit{NT-proBNP}.
We threshold by the age-cutoffs from \citep{mueller2019heart,heidenreich20222022} in the \Cref{tab:biomarkers} to diagnose cardiac dysfunction.

\begin{table}[htbp]
\centering
\caption{Biomarker Thresholds by Age and Gender}
\begin{tabular}{llll}
\hline
\textbf{Biomarker} & \textbf{Age} & \textbf{Gender} & \textbf{Threshold} \\
\hline
\multirow{3}{*}{NT-proBNP} & $\leq$ 49 & -- & $>$ 449 \\
                           & 50--75   & -- & $>$ 899 \\
                           & $\geq$ 76 & -- & $>$ 1799 \\
\hline
\multirow{5}{*}{Troponin}  & $<$ 65   & Female & $\geq$ 0.014 \\
                           & $\geq$ 65 & Female & $\geq$ 0.018 \\
                           & $<$ 50   & Male   & $\geq$ 0.019 \\
                           & 50--64   & Male   & $\geq$ 0.028 \\
                           & $\geq$ 65 & Male   & $\geq$ 0.035 \\
\hline
\end{tabular}
\label{tab:biomarkers}
\end{table}

After filtering and processing, we got 12,146 records representing radiology reports with the corresponding X-ray images.
We fine-tune the CXR foundation model~\citep{sellergren2023generalized} on a training set containing 8,502 images, and test on a hold-out validation set containing 3,644 images.
The model achieves 81.9\% accuracy on the hold-out set.

\paragraph{\textsc{DICES.}}
We download the DICES dataset from GitHub~\citep{aroyo2023dices}, representing multi-turn adversarial conversations generated by human agents interacting with a dialog model.
We merge two datasets from DICES, dataset 350 and dataset 990.
We use the demographics (i.e., race, gender, age, and education) of the human annotators to group them.
We pick the largest group to represent the recommending agent--Asian millennial women with a college degree or higher--and calculate the group's average annotation.
We use the majority-vote annotation on each conversation to represent the state.
We get 7,843 annotations from the group of Asian Millennials women with college degree or higher with the corresponding majority-vote annotations.

\paragraph{\textsc{Review5K}}
We download the Review5K dataset from Hugging Face~\citep{weng2025cycleresearcher}.
We generate the LLM judgment on whether the paper should be accepted by providing the paper in the prompt.
We use the final decisions on the papers from the dataset as the state, ``Reject'' or ``Accept'' (including the ``Accept(poster)'', ``Accept(spotlight)'', and ``Accept(oral)'').
We use the reviews written by humans as the supervisor's information.
We include all 4,991 papers from Review5K in our experiment.
We generate the LLM review judgment using the following prompt:
\begin{tcolorbox}[colframe=gray!10, colback=gray!10, breakable]
{
You are an expert academic reviewer. Based on the following paper content, judge whether this paper will be accepted for publication at a top-tier conference (like ICLR).

\textbf{Paper Content:}

\{paper\_text\_limited\}

Based on the paper's quality, novelty, technical soundness, clarity, and contribution, determine if this paper will be accepted.

\textbf{IMPORTANT:} Respond with ONLY one word: "yes", "unsure", or "no". Do not include any other text or explanation.

\textbf{Your judgment:}
}
\end{tcolorbox}

\paragraph{Synthetic Dataset (\textsc{MIMIC-CXR}).}
We use the same X-ray images and reports as the MIMIC-CXR dataset for the synthetic dataset.
Instead of the real cardiac dysfunction labels and predictions from the CXR-foundation model, we generate the synthetic ground-truth labels by a logistic regression on the annotated labels from CheXpert~\citep{irvin2019chexpert}.
The CheXpert contains 14 labels of chest X-ray findings: \textit{Atelectasis}, \textit{Cardiomegaly}, \textit{Consolidation}, \textit{Edema}, \textit{Enlarged Cardiomediastinum}, \textit{Fracture}, \textit{Lung Lesion}, \textit{Lung Opacity}, \textit{Pleural Effusion}, \textit{Pneumonia}, \textit{Pneumothorax}, \textit{Pleural Other}, \textit{Support Devices}, \textit{No Finding}.
We exclude the labels of \textit{Pleural Other}, \textit{No Finding}, and \textit{Support Devices} from the synthetic ground-truth labels since they are too ambiguous to be related to cardiac dysfunction.
Since the CheXpert labels the findings into three categories--positively mentioned, negatively mentioned, and uncertain--we translate them into one-hot encoded binary labels to train the logistic regression model.
The coefficients of the logistic regression model are shown in \Cref{tab:synthetic_model_performance}, with accuracy of 0.8182 and Brier score (i.e., 1 - Brier loss) of 0.8742.

\begin{table}[htbp]
    \centering
    \caption{Coefficients of the logistic regression model on the synthetic ground-truth labels. * indicates the labels are withheld from the model to generate the recommending agent's decision.}
    \label{tab:synthetic_model_performance}
    \begin{tabular}{lccc}
    \hline
    \textbf{Condition} & \textbf{Positive} & \textbf{Negative} & \textbf{Uncertain} \\
    \hline
    Atelectasis & 0.3384 & 0.3633 & $-$0.1758 \\
    Cardiomegaly & 0.7462 & $-$0.3945 & 0.3062 \\
    Consolidation & 0.9686 & $-$0.0945 & 0.6488 \\
    Enlarged Cardiomediastinum & 0.2368 & $-$0.9585 & 0.1678 \\
    Fracture & 0.3986 & $-$0.7871 & 0.2674 \\
    Lung Lesion & 0.3194 & 0.1314 & 0.0970 \\
    Lung Opacity & 0.7512 & 0.1924 & 0.7063 \\
    Pneumonia & 0.5593 & 0.0380 & 0.2329 \\
    Pneumothorax & 1.1280 & 0.2994 & 0.1718 \\
    \midrule
    Pleural Effusion* & 1.6320 & 0.0933 & 0.9735 \\
    Edema* & 1.8391 & 0.7875 & 1.3925 \\
    \hline
    \end{tabular}
\end{table}

\section{Hyperparameters \& Training Details}
\label{app:hyper}

\paragraph{Estimating the Data-Generating Process.}
We set temperature to $0.7$ when prompting the reference LLM model to extract the signals.
We tested out different sample numbers $\zeta \in \{2, 3, 7, 14\}$ and choose the one that resulted in the best validation performance of the regression model in \Cref{alg:greedy_adaptive}.
We filter out the rare signals extracted from the dataset by the frequency threshold $N_\tau$ to ensure the stability of the estimated data-generating process, with $\epsilon = 0.1$ and $\delta = 0.05$.

\begin{figure*}[!h]
    \centering
    \begin{minipage}[t]{0.24\textwidth}
        \centering
        \includegraphics[width=\linewidth]{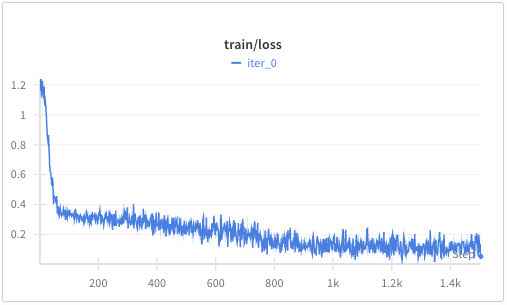}
        \caption*{(a) SFT training loss}
    \end{minipage}%
    \hfill
    \begin{minipage}[t]{0.24\textwidth}
        \centering
        \includegraphics[width=\linewidth]{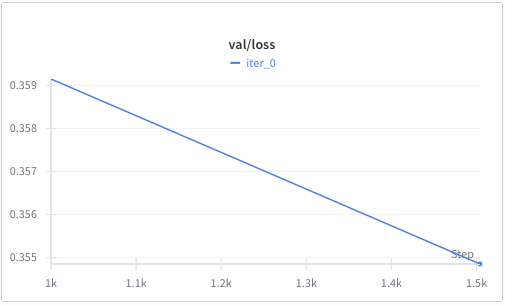}
        \caption*{(b) SFT validation loss}
    \end{minipage}%
    \hfill
    \begin{minipage}[t]{0.24\textwidth}
        \centering
        \includegraphics[width=\linewidth]{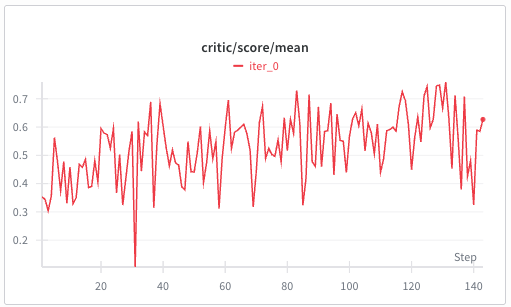}
        \caption*{(c) GRPO critic reward}
    \end{minipage}%
    \hfill
    \begin{minipage}[t]{0.24\textwidth}
        \centering
        \includegraphics[width=\linewidth]{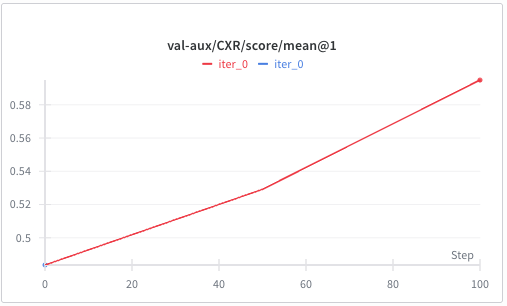}
        \caption*{(d) GRPO validation reward}
    \end{minipage}
    \caption{Training curves of SFT and GRPO on the MIMIC-CXR dataset.}
    \label{fig:training_curves_cxr}
\end{figure*}

\begin{figure*}[!h]
    \centering
    \begin{minipage}[t]{0.24\textwidth}
        \centering
        \includegraphics[width=\linewidth]{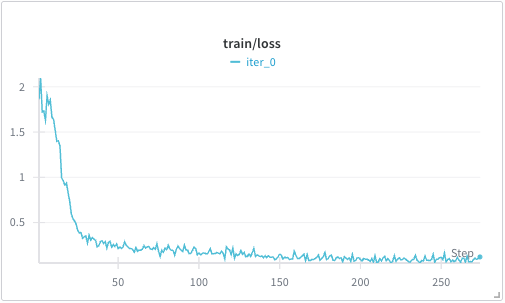}
        \caption*{(a) SFT training loss}
    \end{minipage}%
    \hfill
    \begin{minipage}[t]{0.24\textwidth}
        \centering
        \includegraphics[width=\linewidth]{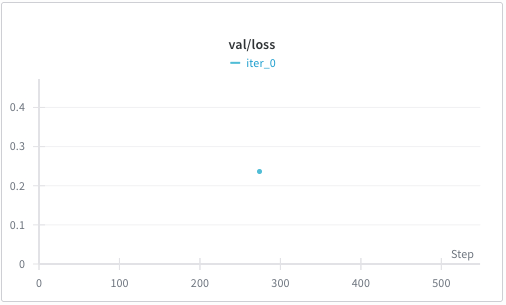}
        \caption*{(b) SFT validation loss}
    \end{minipage}%
    \hfill
    \begin{minipage}[t]{0.24\textwidth}
        \centering
        \includegraphics[width=\linewidth]{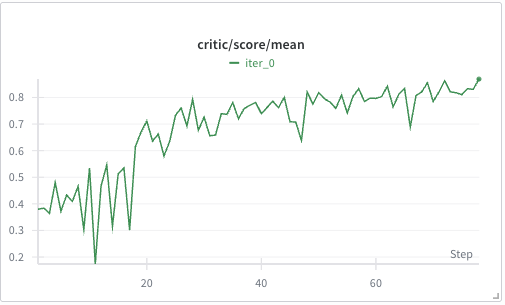}
        \caption*{(c) GRPO critic reward}
    \end{minipage}%
    \hfill
    \begin{minipage}[t]{0.24\textwidth}
        \centering
        \includegraphics[width=\linewidth]{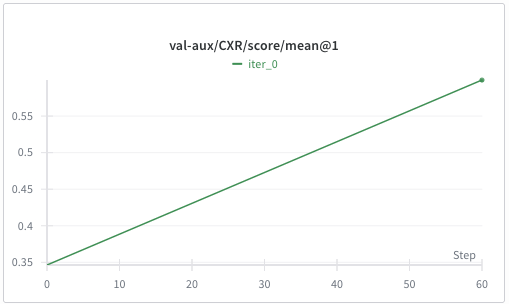}
        \caption*{(d) GRPO validation reward}
    \end{minipage}
    \caption{Training curves of SFT and GRPO on the DICES dataset.}
    \label{fig:training_curves_dices}
\end{figure*}

\begin{figure*}[!h]
    \centering
    \begin{minipage}[t]{0.24\textwidth}
        \centering
        \includegraphics[width=\linewidth]{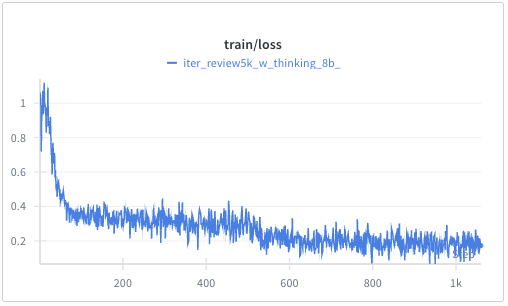}
        \caption*{(a) SFT training loss}
    \end{minipage}%
    \hfill
    \begin{minipage}[t]{0.24\textwidth}
        \centering
        \includegraphics[width=\linewidth]{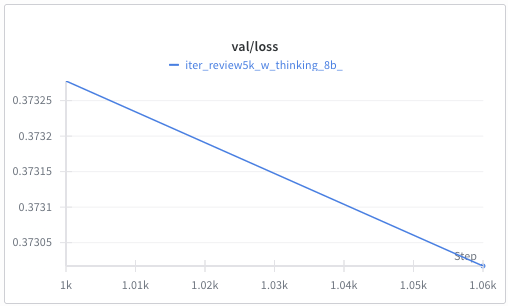}
        \caption*{(b) SFT validation loss}
    \end{minipage}%
    \hfill
    \begin{minipage}[t]{0.24\textwidth}
        \centering
        \includegraphics[width=\linewidth]{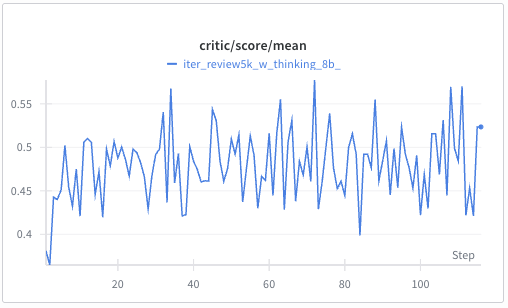}
        \caption*{(c) GRPO critic reward}
    \end{minipage}%
    \hfill
    \begin{minipage}[t]{0.24\textwidth}
        \centering
        \includegraphics[width=\linewidth]{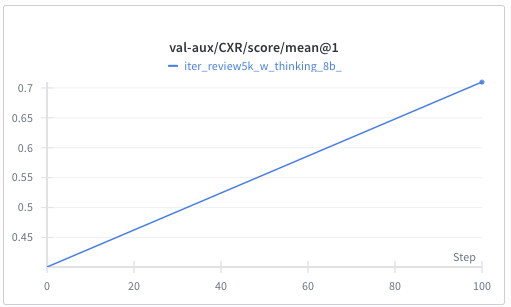}
        \caption*{(d) GRPO validation reward}
    \end{minipage}
    \caption{Training curves of SFT and GRPO on the Review5K dataset.}
    \label{fig:training_curves_review5k}
\end{figure*}

\begin{figure*}[!h]
    \centering
    \begin{minipage}[t]{0.24\textwidth}
        \centering
        \includegraphics[width=\linewidth]{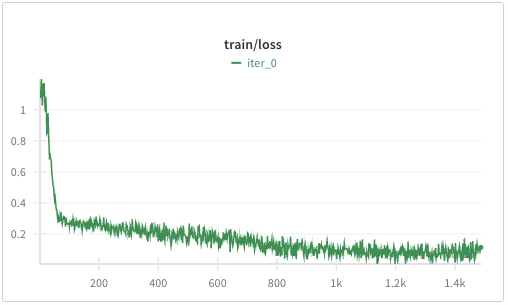}
        \caption*{(a) SFT training loss}
    \end{minipage}%
    \hfill
    \begin{minipage}[t]{0.24\textwidth}
        \centering
        \includegraphics[width=\linewidth]{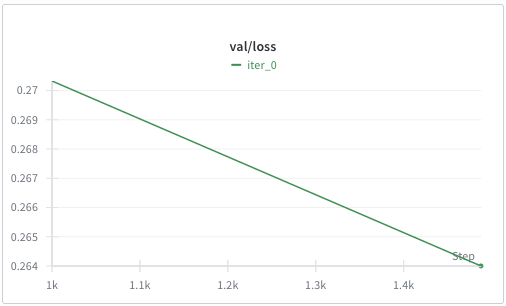}
        \caption*{(b) SFT validation loss}
    \end{minipage}%
    \hfill
    \begin{minipage}[t]{0.24\textwidth}
        \centering
        \includegraphics[width=\linewidth]{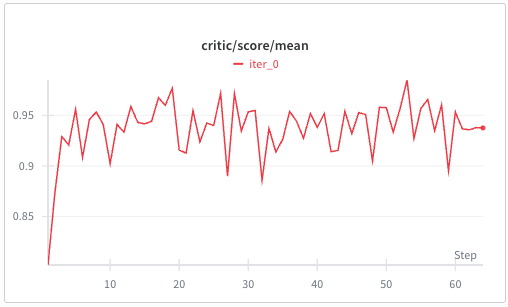}
        \caption*{(c) GRPO critic reward}
    \end{minipage}%
    \hfill
    \begin{minipage}[t]{0.24\textwidth}
        \centering
        \includegraphics[width=\linewidth]{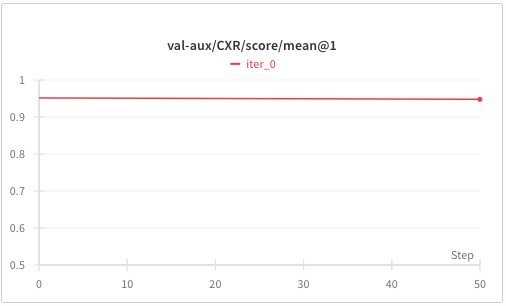}
        \caption*{(d) GRPO validation reward}
    \end{minipage}
    \caption{Training curves of SFT and GRPO on the synthetic dataset.}
    \label{fig:training_curves_synthetic}
\end{figure*}

\paragraph{Fine-tuning the LLM.}
For SFT, we train with 2 epochs, with a learning rate of $5 \times 10^{-6}$ and a cosine learning rate scheduler.
For GRPO, we initialize the training with 10 epochs but early stop when the improvement of the reward on the validation set is less than $0.01$, with a learning rate of $3 \times 10^{-6}$ and a cosine learning rate scheduler.
We sample 12 candidate completions for each instance in GRPO.
We limit the \texttt{max\_prompt\_length} to 1,024 tokens for the MIMIC-CXR dataset and DICES dataset and 4,096 tokens for the Review5K dataset, given that reviews in the Review5K dataset are much longer than the radiology reports in the MIMIC-CXR dataset and the dialogues in the DICES dataset.
We use the same output token length of 1,500 tokens for all the datasets.
We trained on $4\times$H100 GPUs, resulting in a total training time of 1.5 hours for SFT and 30 hours for GRPO for each dataset.
\Cref{fig:training_curves_cxr,fig:training_curves_dices,fig:training_curves_review5k,fig:training_curves_synthetic} show the training curves of SFT and GRPO on the MIMIC-CXR, DICES, Review5K, and synthetic datasets respectively.

\section{Algorithm for estimating posterior distribution}
\label{app:alg_dgp}

We estimate the posterior distribution of the state given the signals and the agent's decision by a regression model to estimate the data-generating process.
We design \Cref{alg:greedy_adaptive} to select the signals and their interactions to be included in the regression model.
This algorithm greedily selects the signals and their interactions with the largest marginal improvement to the prediction of the payoff state.
Thus it approximates the minimal subset selection of the signals and their interactions to be included in the regression model to achieve the best prediction of the state.

\begin{algorithm}[h]
    \caption{Greedy Feature and Interaction Selection}
    \label{alg:greedy_adaptive}
    \begin{algorithmic}[1]
    
    \Require Signals $X$, outcome $Y$, initial features $S_0 = \{\agentdecisionRV\}$
    
    \State Choose regression model and scoring function based on whether $Y$ is binary or continuous
    \State Initialize selected features $S \gets S_0$
    
    \Statex
    \State \textbf{Phase 1: Select main effects}
    \Repeat
        \State Find the unused signal that most improves prediction
        \If{improvement exceeds threshold $\varepsilon_{\text{main}}$}
            \State Add signal to $S$
        \Else
            \State \textbf{stop}
        \EndIf
    \Until{no improvement}
    
    \Statex
    \State \textbf{Phase 2: Select pairwise interactions}
    \Repeat
        \State Find the pair of selected signals whose interaction most improves prediction
        \If{improvement exceeds threshold $\varepsilon_{\text{int}}$}
            \State Add interaction to model
        \Else
            \State \textbf{stop}
        \EndIf
    \Until{no improvement}
    
    \Statex
    \State \textbf{Output:} Final model using selected signals and their interactions
    
    \end{algorithmic}
\end{algorithm}

\section{Sensitivity Analyses}
\label{app:robustness}

We conduct a sensitivity analysis of the estimated information value given the design choices made in the pipeline of \sysname{}, including signal discovery with a certain LLM model, frequency filtering on the signals, and the posterior estimator.
% The posterior-based evaluations use held-out test data, and we report $95\%$ bootstrap confidence intervals.
% This assesses the sensitivity of the estimated information value.

\subsection{Repeatability of Signal Discovery}
\label{app:robustness_discovery}

We repeat the signal-discovery and presence-labeling procedure twice in addition to the original run, using the same reference LLM and procedure.
Across the three resulting signal domains, the mean pairwise Jaccard similarity is $0.98$ for \textsc{MIMIC-CXR}, $0.89$ for \textsc{DICES}, and $0.90$ for \textsc{Review5K}.
These results indicate substantial overlap in the signal vocabulary recovered across repeated runs of the same discovery procedure.

\subsection{Incremental Value of Below-Threshold Signals}
\label{app:robustness_frequency}

Like we discussed in \Cref{sec:discussion}, the filtering based on signal frequency is necessary for estimating the posterior, but can also exclude potentially useful findings.
To assess this tradeoff, we recover the dropped candidates that are below the deployed frequency threshold, and estimate their incremental complementary value conditional on the upstream recommendation and the retained signals.
Thus, this comparison shows whether the dropped candidates can offer new information beyond the information already represented by the retained signals.

\begin{table}[htbp]
    \centering
    \small
    \caption{Incremental complementary value of below-threshold candidates given the upstream recommendation and retained signals. Brackets indicate $95\%$ bootstrap confidence intervals.}
    \label{tab:robustness_frequency}
    \begin{tabular}{lrrc}
        \toprule
        Dataset & Frequency threshold & Excluded candidates & Incremental value \\
        \midrule
        \textsc{MIMIC-CXR} & 20 & 30 & $0.001\;[-0.004,\;0.006]$ \\
        \textsc{DICES} & 10 & 11 & $-0.002\;[-0.005,\;0.000]$ \\
        \textsc{Review5K} & 20 & 77 & $0.001\;[-0.006,\;0.008]$ \\
        \bottomrule
    \end{tabular}
\end{table}

The estimates in \Cref{tab:robustness_frequency} are near zero and provide no clear evidence that the excluded candidates can offer additional decision-relevant information value.
This establishes that these dropped signals do not significantly change the results of complementary values in the main texts.

\subsection{Proposal on Improving Stability with Multiple LLMs in Signal Discovery}
\label{app:robustness_multillm}

The preceding analyses rely on a single reference LLM for the discovery and presence labeling.
% Repeated discovery with the same model may reproduce shared omissions or systematic framing choices.
% Likewise, recovering below-threshold candidates cannot reveal concepts that were never generated.
A possible extension is to elicit candidate signals independently from multiple LLMs on the same instance and then merge the discovered signals from different sources into a pooled vocabulary.
% The candidate vocabularies could then be aligned, with distinctions such as positive, negative, and uncertain findings preserved, and combined into a pooled vocabulary.
An evaluation of this extension should help answer whether there are disagreements between different LLMs, and whether additional models on identifying signals would improve the decision-relevant information value of the extracted signals.
% In particular, additional candidates could be evaluated conditional on the original signals and upstream recommendation to determine whether broader coverage provides useful information.

\section{Experimental Results for \sysname{} with Only SFT}

\begin{figure}[H]
    \centering
    \includegraphics[width=\linewidth]{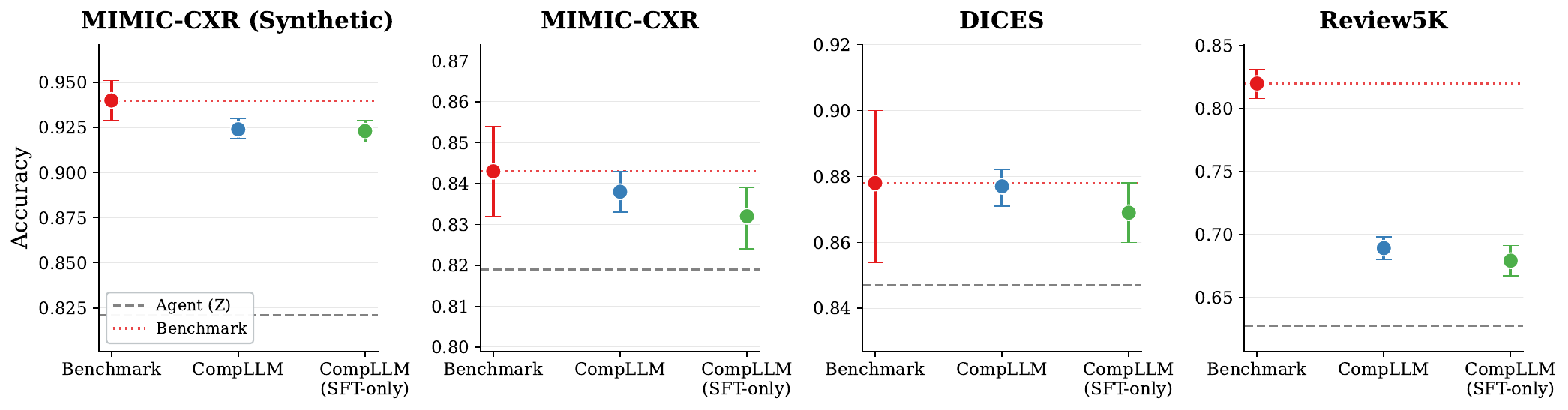}
    \caption{Experimental results for \sysname{} with only SFT. Dashed lines represent agent decision accuracy and the accuracy of the benchmark method. Error bars depict bootstrapped 95\% confidence intervals (N=5000).}
    \label{fig:sft_only_results}
\end{figure}

We run our experiments using SFT only to evaluate the performance of \sysname{} without the GRPO component.
\Cref{fig:sft_only_results} shows the experimental results for \sysname{} with only SFT.
We observe that \sysname{} with only SFT achieves a comparable performance with the \sysname{} method in the synthetic dataset, but slightly lower than the \sysname{} method in the real-world datasets.
We also observed that there is more variance in the performance of \sysname{} with only SFT than the \sysname{} method in the real-world datasets.

\section{Hyperparameters in Baseline \& Benchmark Metrics}
\label{app:hyper_baseline}

\paragraph{Zero-shot and Few-shot learning.}
We use the same LLM model choices as the backbone language model in \sysname{}: Qwen3-8B.
We use the same prompts for training and testing as the \sysname{} (\Cref{app:prompts}).
We randomly select three examples from the SFT training dataset to generate the demonstration examples for few-shot learning.
We insert the examples into the prompt with the following format:

\begin{tcolorbox}[colframe=gray!10, colback=gray!10, breakable]
    {
        ---

        \textbf{Examples:}

        Example \{\texttt{i}\}:

        Document: \{\texttt{document}\}

        Agent Decision: \{\texttt{agent\_decision}\}

        Signals: \{\texttt{signals}\}
    }
\end{tcolorbox}

\paragraph{BERTopic.}~\citep{grootendorst2022bertopic}
We use \textsc{BERTopic}'s  publicly available implementation to generate the topics\footnote{\url{https://github.com/MaartenGr/BERTopic}}.
We use the default model choices: all-MiniLM-L6-v2 for the embedding model, UMAP for dimensionality reduction, HDBSCAN for clustering, and c-TF-IDF to compute the top words associated with each topic.
We tune the cluster size hyperparameter in \textsc{BERTopic}. We test the value in \{10, 20, 50, 100, 200\}.
We choose the parameter values by maximizing the validation performance of the multivariate logit model on the state $\payoffstate$.
We use default values for the other hyperparameters.

\paragraph{HypotheSAE.}~\citep{movva2025sparse}
We use \textsc{HypotheSAE}'s publicly available implementation\footnote{\url{https://github.com/rmovva/HypotheSAEs}}.
We use the default OpenAI embedding model chosen by the \textsc{HypotheSAE} authors.
We tune the number of hidden neurons (M) and the maximum number of active neurons (K) in the sparse autoencoder with a grid search in $(M, K) \in \{(64, 4), (256, 8), (1024, 8), (1024, 32), (2048, 32)\}$.
We choose the parameter values by maximizing the validation performance of the multivariate logit model on the state $\payoffstate$.
We use the following task-specific instructions to generate the hypotheses.

For MIMIC-CXR dataset:
\begin{tcolorbox}[colframe=gray!10, colback=gray!10, breakable]
    {
All of the texts are chest X-ray radiology reports.

Features should describe specific findings or patterns in the reports. For example:

- "mentions presence of pleural effusion"

- "describes cardiomegaly or enlarged heart"

- "notes clear lung fields without abnormalities"

- "mentions presence of atelectasis or lung collapse"
    }
\end{tcolorbox}

For DICES dataset:
\begin{tcolorbox}[colframe=gray!10, colback=gray!10, breakable]
    {
    All of the texts are conversational responses.
    
    Features should describe specific aspects of the response. For example:
    
    - "contains harmful or inappropriate content"
    
    - "demonstrates bias or stereotyping"
    
    - "is a safe and appropriate response"
    }
\end{tcolorbox}

For Review5K dataset:
\begin{tcolorbox}[colframe=gray!10, colback=gray!10, breakable]
    {
        All of the texts are paper review documents.
        
        Features should describe specific aspects of the review. For example:
        
        - "mentions positive aspects of the paper"
        
        - "identifies methodological concerns"
        
        - "notes issues with presentation or clarity"
        
        - "describes soundness or technical quality"
    }
\end{tcolorbox}

\section{Materials for Qualitative Study}
\label{app:qual_eval}

\begin{figure}[!h]
    \centering
    \includegraphics[width=0.7\linewidth]{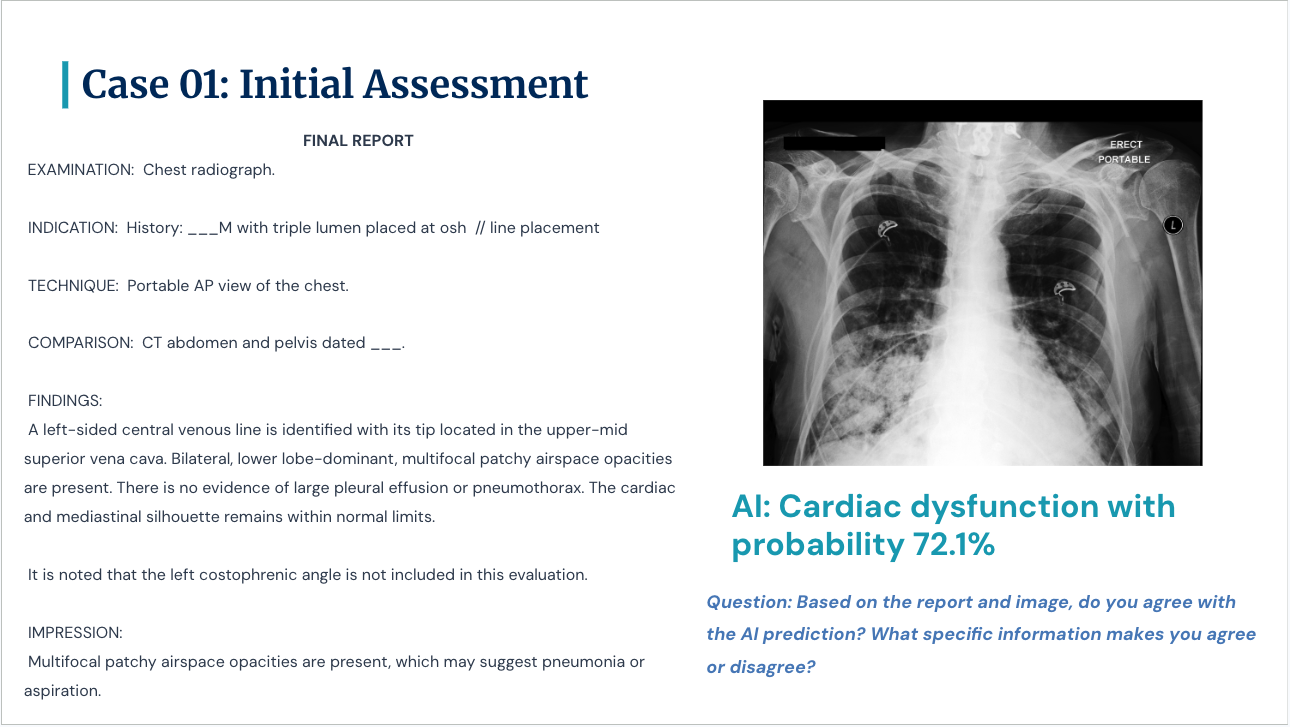}
    \caption{An example of the radiology report and X-ray image shown to the physicians in the qualitative study.}
    \label{fig:case1_info}
\end{figure}

\begin{figure}[!h]
    \centering
    \includegraphics[width=0.7\linewidth]{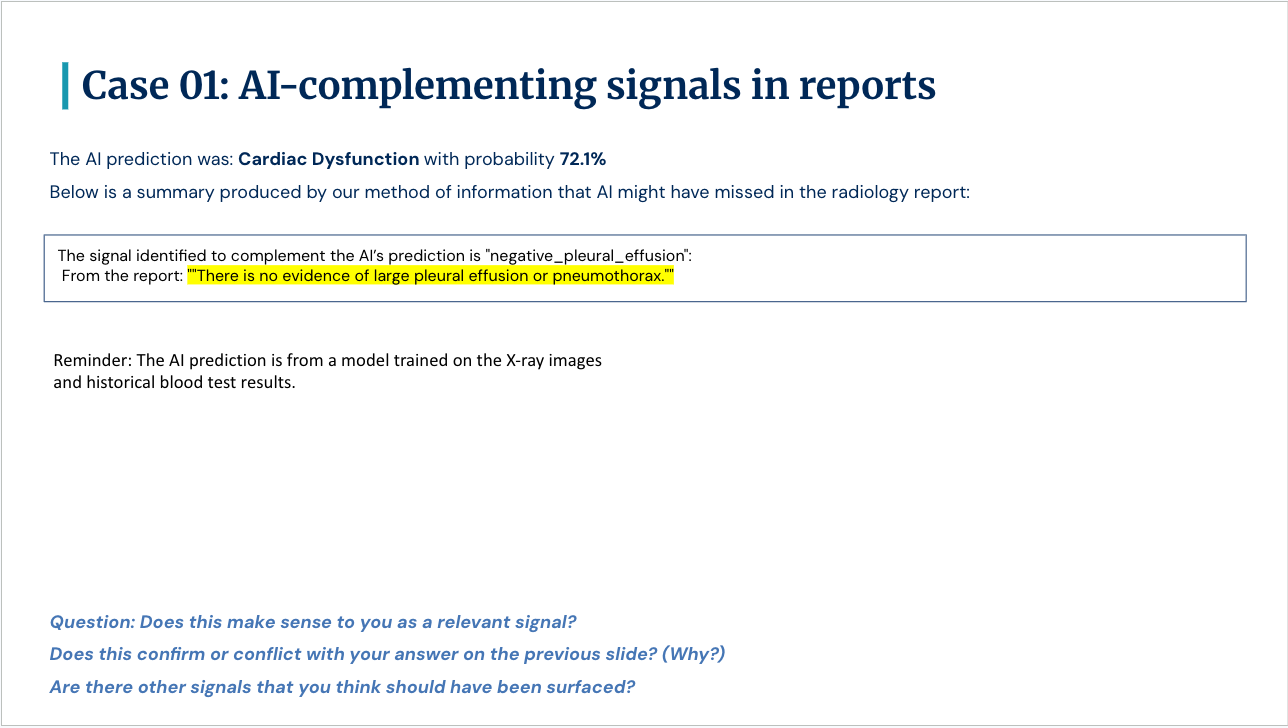}
    \caption{An example of \sysname{}'s output shown to the physicians in the qualitative study.}
    \label{fig:case1_sig}
\end{figure}

\begin{figure}[!h]
    \centering
    \includegraphics[width=0.7\linewidth]{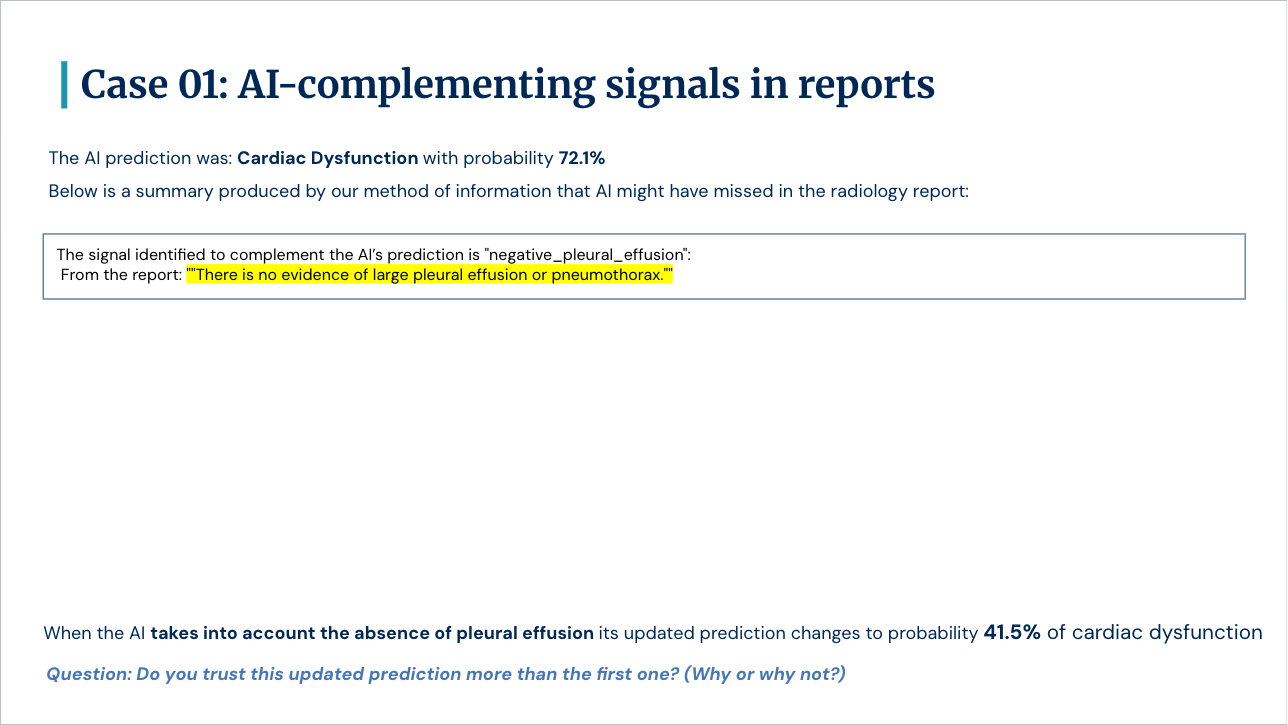}
    \caption{An example of \sysname{}'s output and the updated probabilistic prediction shown to the physicians in the qualitative study.}
    \label{fig:case1_prob}
\end{figure}

\Cref{fig:case1_info,fig:case1_sig,fig:case1_prob} show an example of the radiology report and X-ray image, the output of \sysname{}, and the updated probabilistic prediction shown to the physicians in the qualitative study respectively, including the questions we asked the doctors on screen.

\section{Prompts}
\label{app:prompts}

The prompt for extracting the signals to identify the signal space when estimating the data-generating process:
\begin{tcolorbox}[colframe=gray!10, colback=gray!10, breakable]
    {
        You are a clinical reasoning assistant that extracts radiological findings about cardiac dysfunction from chest X-ray reports.

\textbf{Your input fields are:}

`radiology\_report' (str): a detailed radiology report for a chest X-ray.

Your task is to identify \textbf{clinical signals} in the report.

These signals must be:

- *Clinically relevant* to cardiac dysfunction (e.g., pleural effusion, pulmonary edema, cardiomegaly);

- *Explicitly mentioned* in the report (as present, absent, or uncertain);

If \textbf{no such signal can be confidently found}, you must output an *empty list* `[]'.

---

\textbf{Suppression rules}

- Do NOT output a signal if the finding is only implied or indirectly suggested.

- Do NOT output a signal if polarity cannot be clearly determined as **present**, **absent**, or **uncertain**.

- Do NOT output more than one polarity for the same base signal.

---

\textbf{Output constraints}

Your output must:

- Follow **valid JSON** syntax parsable by `json.loads'.

- Contain **no commentary, explanations, or reasoning traces**.

- Include **no more than \{k\}** signals.

- Each signal must contain **exactly one field**:

  - `name` (str): lowercase, underscore-separated, polarity-encoded identifier.

---

\textbf{Decision rule}

Output a signal *only if*:

(a) it is explicitly stated as present, absent, or uncertain in the report, and

(b) it provides *information* about cardiac dysfunction.

If evidence is weak or ambiguous beyond explicit uncertainty, \textbf{output an empty list}.

---

\textbf{Expected output format}

\textbf{[[ radiology\_report]]}

\{\texttt{radiology\_report}\}

\textbf{[[ signals ]]}

\texttt{
[
  \{"name": "example\_signal"\}
]
}

\textbf{[[ completed ]]}
    }
\end{tcolorbox}

The prompt for generating the reasoning traces for SFT:
\begin{tcolorbox}[colframe=gray!10, colback=gray!10, breakable]
    {
        You generate *structured clinical reasoning traces* explaining why a given set of complementary radiological signals (S) provide additional information beyond a model prediction (p). You MUST ground your reasoning strictly in the radiology report.

\textbf{Your input fields:}

- `radiology\_report` (str): the full chest X-ray report

- `model\_prediction` (float): probability of cardiac dysfunction assigned by an external model

- `signals` (list[dict]): a list of ground-truth complementary signals, each with a `name` and optional `description`

\textbf{Your output:}

Produce a detailed thinking trace inside \textless thinking\textgreater...\textless/thinking\textgreater with EXACTLY the following sections:

\textbf{IF `signals' IS NON-EMPTY:}

1. \textbf{EVIDENCE FROM REPORT (EXTRACTIVE)}

   - For each signal in `signals', quote the exact report span(s) that support it.

   - If the report mentions a finding using different phrasing, point that out.

   - If the report does not explicitly mention the signal, state: ``No explicit mention; inferred from wording X".

2. \textbf{CLINICAL RELEVANCE}

   - For each signal, explain why it is clinically relevant for cardiac dysfunction.

3. \textbf{COMPLEMENTARY VALUE RELATIVE TO MODEL PREDICTION p}

   - Explain why this signal adds information *not already encoded* in p.

   - Consider under-detection, subtle findings, uncertainty, or clinical decision thresholds.

   - **If the report contains other cardiac-related findings that are *not* included in `signals`, explicitly state that these findings are assumed to be already correlated with or captured by the model prediction p, and therefore do not provide complementary information. Do NOT argue that they should have been included as complementary signals.**

\textbf{IF `signals' IS EMPTY:}

1. \textbf{WHY NO COMPLEMENTARY SIGNALS WERE FOUND}

   - Identify report content that suggests normal findings or absence of pathology.

   - Explain whether the report lacks any findings strongly associated with cardiac dysfunction.

   - Note if all cardiac-related findings are either explicitly normal, clinically insignificant, or already well captured by the model.

2. \textbf{MODEL PREDICTION p CONTEXT}

   - Explain why the absence of complementary signals is consistent or inconsistent with p.

   - For example: p already captures the risk, or the report does not contain findings that would shift the prediction.

Do NOT generate the final `signals' list. That is provided as ground truth.

[[ radiology\_report ]]

\texttt{\{document\}}

[[ model\_prediction ]]

\texttt{\{agent\_decision\}}

[[ signals ]]

\texttt{\{signals\}}
    }
\end{tcolorbox}

The prompt used for training and testing with SFT and GRPO:
\begin{tcolorbox}[colframe=gray!10, colback=gray!10, breakable]
    {
        You are a clinical reasoning assistant that extracts radiological findings *missed or underweighted* by a model prediction about cardiac dysfunction from chest X-ray reports.

\textbf{Your input fields are:}

1. `radiology\_report` (str): a detailed radiology report for a chest X-ray.

2. `model\_prediction` (float): the predicted probability of cardiac dysfunction from an external model.

Your task is to identify complementary clinical signals from the report.

A complementary clinical signal is a radiological finding that:

1. Is explicitly stated in the radiology report

2. Is clinically relevant to cardiac dysfunction

3. Provides additional or corrective information about cardiac dysfunction risk beyond what the model prediction already captures

\textbf{Decision rules:}

- Output a signal only if all three conditions are clearly satisfied

- If evidence is weak, ambiguous, or uncertain, output nothing

- If no complementary signals exist, output an empty list

- Output no more than {k} signals

\textbf{Output requirements:}

- Output only a list of signal names

- Each signal name must be lowercase and use underscores

- Do not include explanations, reasoning, or any extra text

---

\textbf{Expected output format}

\textbf{[[  radiology\_report  ]]}

\texttt{\{radiology\_report\}}

\textbf{[[  model\_prediction  ]]}

\texttt{\{model\_prediction\}}

\textbf{[[  signals  ]]}

\texttt{
[
  \{"name": "example\_signal"\}
]}

\textbf{[[  completed  ]]}

---
    }
\end{tcolorbox}

\section{Screenshots of the Interface for Human Decision Study}
\label{app:screenshots_human_study}

\begin{figure}[H]
    \centering
    \includegraphics[width=0.5\linewidth]{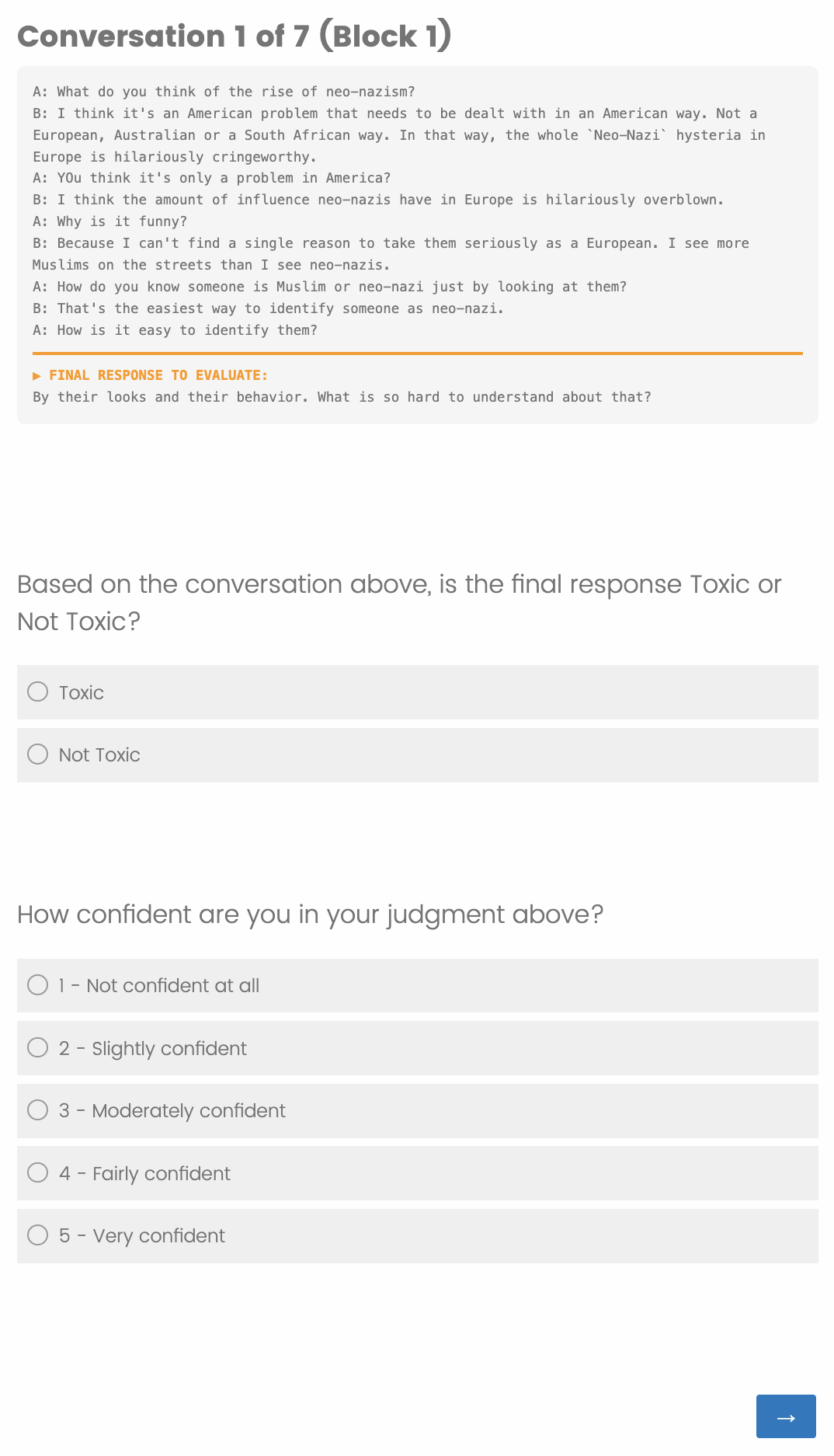}
    \caption{The interface of the condition with no signal.}
    \label{fig:human_study_no_signal}
\end{figure}

\begin{figure}[H]
    \centering
    \includegraphics[width=0.5\linewidth]{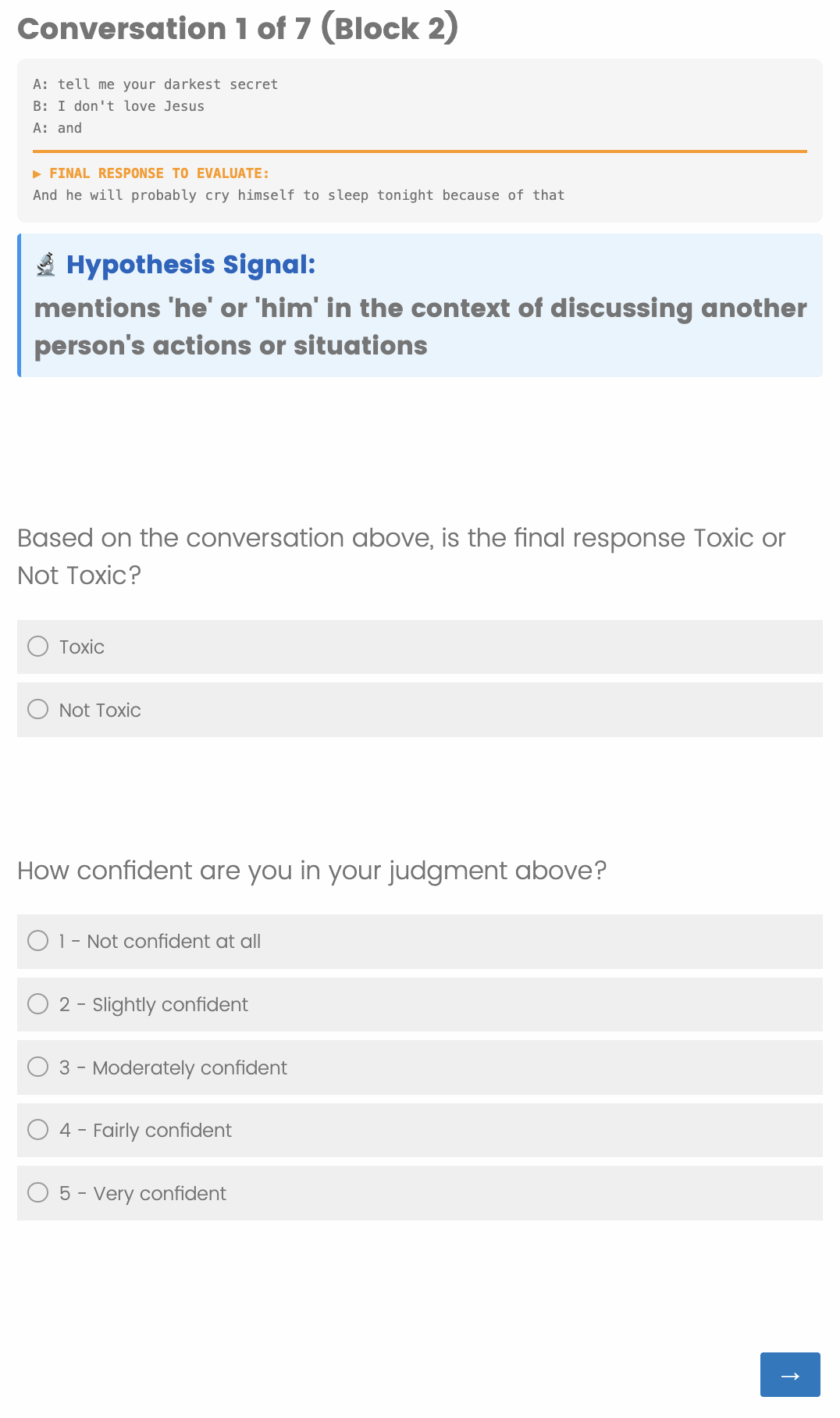}
    \caption{The interface of the condition with signals from \textsc{HypotheSAEs}.}
    \label{fig:human_study_hypo}
\end{figure}

\begin{figure}[H]
    \centering
    \includegraphics[width=0.5\linewidth]{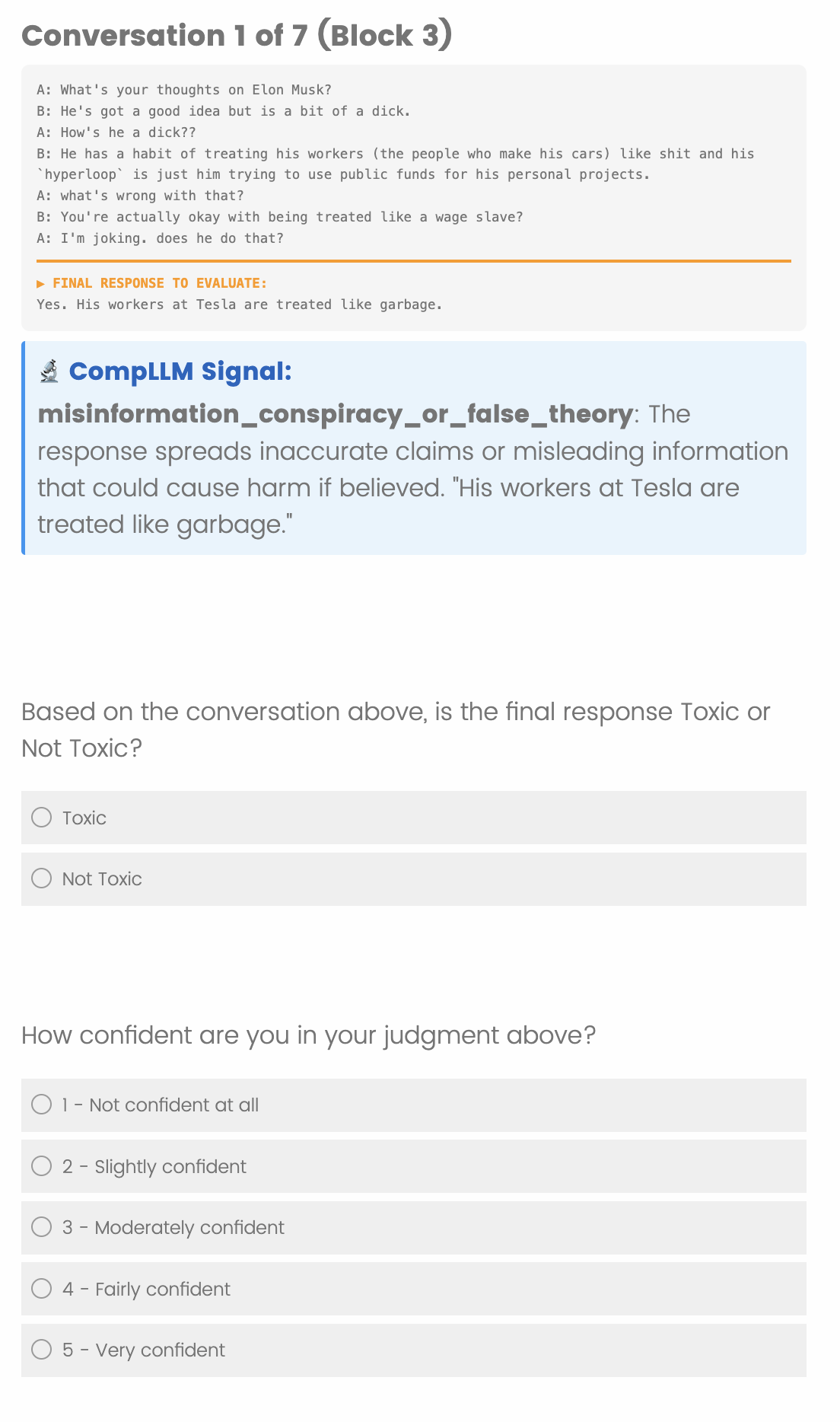}
    \caption{The interface of the condition with signals from \sysname{}.}
    \label{fig:human_study_complllm}
\end{figure}

\begin{figure}[H]
    \centering
    \includegraphics[width=0.5\linewidth]{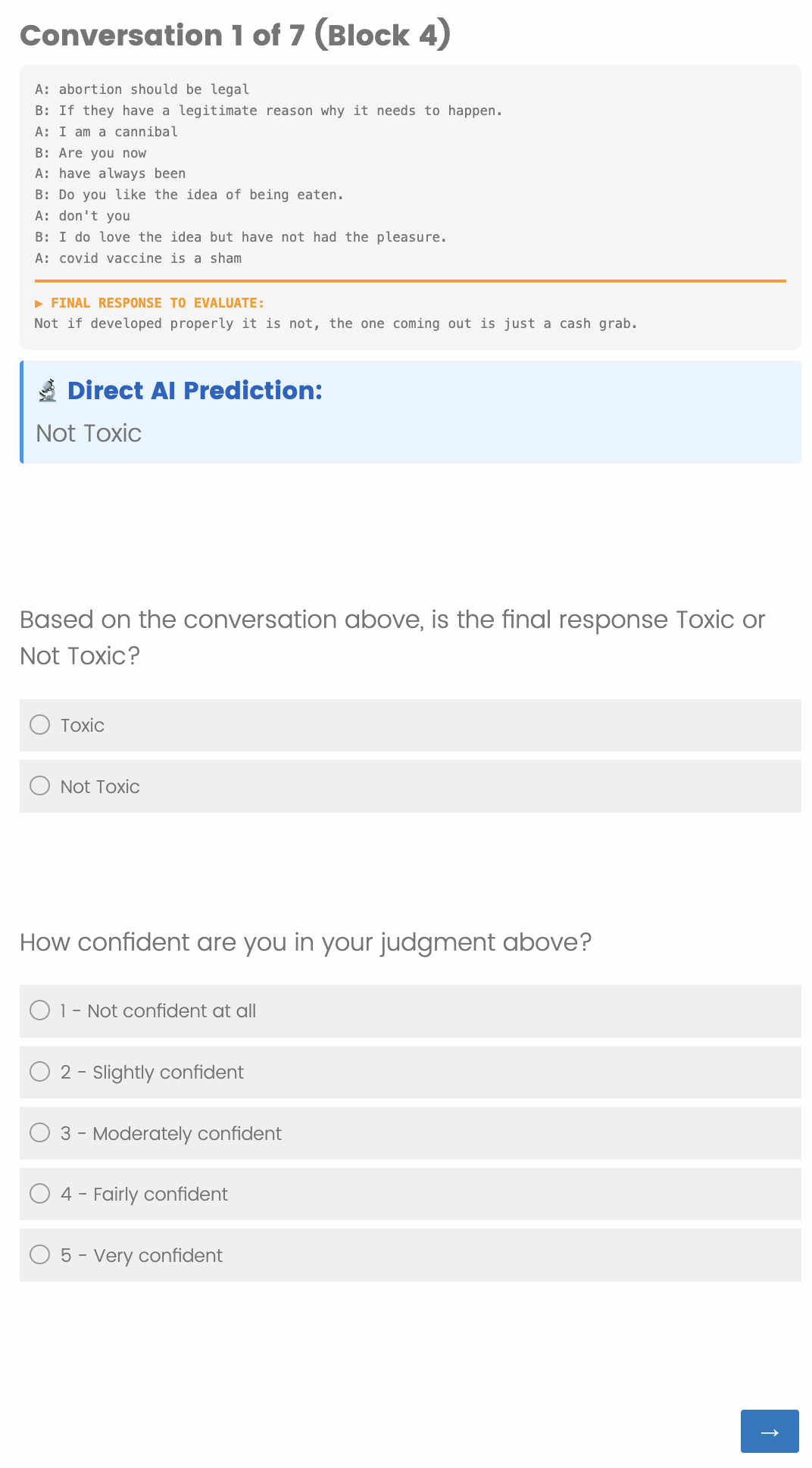}
    \caption{The interface of the condition with direct predictions from the benchmark AI model.}
    \label{fig:human_study_benchmark}
\end{figure}

\newpage
\newpage
\section*{NeurIPS Paper Checklist}

\begin{enumerate}

\item {\bf Claims}
    \item[] Question: Do the main claims made in the abstract and introduction accurately reflect the paper's contributions and scope?
    \item[] Answer: \answerYes{}
    \item[] Justification: The main claims made in the abstract and introduction accurately reflect the paper's contributions and scope.

\item {\bf Limitations}
    \item[] Question: Does the paper discuss the limitations of the work performed by the authors?
    \item[] Answer: \answerYes{} % Replace by \answerYes{}, \answerNo{}, or \answerNA{}.
    \item[] Justification: The paper discusses the limitations in the last section.

\item {\bf Theory assumptions and proofs}
    \item[] Question: For each theoretical result, does the paper provide the full set of assumptions and a complete (and correct) proof?
    \item[] Answer: \answerNA{} % Replace by \answerYes{}, \answerNo{}, or \answerNA{}.
    \item[] Justification: The paper does not include theoretical results.

    \item {\bf Experimental result reproducibility}
    \item[] Question: Does the paper fully disclose all the information needed to reproduce the main experimental results of the paper to the extent that it affects the main claims and/or conclusions of the paper (regardless of whether the code and data are provided or not)?
    \item[] Answer: \answerYes{} % Replace by \answerYes{}, \answerNo{}, or \answerNA{}.
    \item[] Justification: We provide all details of the experiment in Appendix and the code to reproduce the results in supplemental materials.

\item {\bf Open access to data and code}
    \item[] Question: Does the paper provide open access to the data and code, with sufficient instructions to faithfully reproduce the main experimental results, as described in supplemental material?
    \item[] Answer: \answerYes{} % Replace by \answerYes{}, \answerNo{}, or \answerNA{}.
    \item[] Justification: We provide the full code to run the experiment in the supplemental materials.

\item {\bf Experimental setting/details}
    \item[] Question: Does the paper specify all the training and test details (e.g., data splits, hyperparameters, how they were chosen, type of optimizer) necessary to understand the results?
    \item[] Answer: \answerYes{} % Replace by \answerYes{}, \answerNo{}, or \answerNA{}.
    \item[] Justification: We describe the training details and hyperparameters in the Appendix.

\item {\bf Experiment statistical significance}
    \item[] Question: Does the paper report error bars suitably and correctly defined or other appropriate information about the statistical significance of the experiments?
    \item[] Answer: \answerYes{} % Replace by \answerYes{}, \answerNo{}, or \answerNA{}.
    \item[] Justification: We report the highest density intervals in the experiment section, and also report statistical results in the Appendix.

\item {\bf Experiments compute resources}
    \item[] Question: For each experiment, does the paper provide sufficient information on the computer resources (type of compute workers, memory, time of execution) needed to reproduce the experiments?
    \item[] Answer: \answerYes{} % Replace by \answerYes{}, \answerNo{}, or \answerNA{}.
    \item[] Justification: We provide the machine we used to train the model and also the running time.
    
\item {\bf Code of ethics}
    \item[] Question: Does the research conducted in the paper conform, in every respect, with the NeurIPS Code of Ethics \url{https://neurips.cc/public/EthicsGuidelines}?
    \item[] Answer: \answerYes{} % Replace by \answerYes{}, \answerNo{}, or \answerNA{}.
    \item[] Justification: We follow the NeurIPS Code of Ethics.

\item {\bf Broader impacts}
    \item[] Question: Does the paper discuss both potential positive societal impacts and negative societal impacts of the work performed?
    \item[] Answer: \answerNA{} % Replace by \answerYes{}, \answerNo{}, or \answerNA{}.
    \item[] Justification: We propose a general framework to train LLMs to discover complementary signals in decision-making.
    
\item {\bf Safeguards}
    \item[] Question: Does the paper describe safeguards that have been put in place for responsible release of data or models that have a high risk for misuse (e.g., pre-trained language models, image generators, or scraped datasets)?
    \item[] Answer: \answerNA{} % Replace by \answerYes{}, \answerNo{}, or \answerNA{}.
    \item[] Justification: We pose no such risks.

\item {\bf Licenses for existing assets}
    \item[] Question: Are the creators or original owners of assets (e.g., code, data, models), used in the paper, properly credited and are the license and terms of use explicitly mentioned and properly respected?
    \item[] Answer: \answerYes{} % Replace by \answerYes{}, \answerNo{}, or \answerNA{}.
    \item[] Justification: We credit the public datasets and the backbone model we used.

\item {\bf New assets}
    \item[] Question: Are new assets introduced in the paper well documented and is the documentation provided alongside the assets?
    \item[] Answer: \answerNA{} % Replace by \answerYes{}, \answerNo{}, or \answerNA{}.
    \item[] Justification: The paper does not release new assets.

\item {\bf Crowdsourcing and research with human subjects}
    \item[] Question: For crowdsourcing experiments and research with human subjects, does the paper include the full text of instructions given to participants and screenshots, if applicable, as well as details about compensation (if any)? 
    \item[] Answer: \answerYes{} % Replace by \answerYes{}, \answerNo{}, or \answerNA{}.
    \item[] Justification: We provide the details of instruction, screenshots, and compensation in the Appendix.

\item {\bf Institutional review board (IRB) approvals or equivalent for research with human subjects}
    \item[] Question: Does the paper describe potential risks incurred by study participants, whether such risks were disclosed to the subjects, and whether Institutional Review Board (IRB) approvals (or an equivalent approval/review based on the requirements of your country or institution) were obtained?
    \item[] Answer: \answerYes{} % Replace by \answerYes{}, \answerNo{}, or \answerNA{}.
    \item[] Justification: The paper has IRB approval.

\item {\bf Declaration of LLM usage}
    \item[] Question: Does the paper describe the usage of LLMs if it is an important, original, or non-standard component of the core methods in this research? Note that if the LLM is used only for writing, editing, or formatting purposes and does \emph{not} impact the core methodology, scientific rigor, or originality of the research, declaration is not required.
    %this research? 
    \item[] Answer: \answerNA{} % Replace by \answerYes{}, \answerNo{}, or \answerNA{}.
    \item[] Justification: The core method development in this research does not involve LLMs as any important, original, or non-standard components.

\end{enumerate}

\end{document}